\documentclass[a4paper,10pt] {article}

\usepackage{amsfonts,amsmath,amssymb,amsthm}
\usepackage[final]{graphicx}
\usepackage{fancyhdr,sectsty}
\usepackage{hyperref,microtype,makeidx}
\usepackage[nottoc]{tocbibind}
\usepackage{textcomp,float}
\usepackage[all]{xy}
\usepackage{float}
\usepackage{color,pstricks}
\usepackage{fullpage}
\usepackage{setspace}
\usepackage{caption}
\usepackage{subcaption}
\usepackage[authoryear]{natbib}

\usepackage{hyperref}
\usepackage{color}
\usepackage{soul}
\usepackage{caption}
\usepackage{subcaption}
\usepackage{amsmath}
\usepackage{amssymb}
\usepackage{algorithm,algorithmic}
\usepackage{paralist}
\usepackage{multirow}

\theoremstyle{plain}

\title{Local and global gestalt laws: A neurally based spectral approach}
\author{Marta Favali\footnote{M. Favali,
		Center of Mathematics, CNRS - EHESS, Paris, France. marta.favali@ehess.fr}, Giovanna Citti\footnote{ G. Citti, Dipartimento di Matematica, Universit\`{a} di Bologna, Bologna, Italy. giovanna.citti@unibo.it} and Alessandro Sarti\footnote{A. Sarti, Center of Mathematics, CNRS - EHESS, Paris, France. alessandro.sarti@ehess.fr}} 
\date{}

\begin{document}
	\maketitle
	
	\begin{abstract}
		
		A mathematical model of figure-ground articulation is presented, which takes into account both local and global gestalt laws and is compatible with the functional architecture of the primary visual cortex (V1). 
The local gestalt law of good continuation is described by means of suitable connectivity kernels, that are derived from Lie group theory, and quantitatively compared with long range connectivity in V1. 
		Global gestalt constraints are then introduced in terms of spectral analysis of connectivity matrix derived from these kernels. This analysis performs grouping of local features and individuates perceptual units with the highest saliency. Numerical simulations are performed and results are obtained applying the technique to a number of stimuli.\\
		\textbf{keywords:} Mathematical modelling, Quantitative gestalt, Figure-Ground Segmentation, Perceptual Grouping, Neural models, Cortical architecture.		
	\end{abstract}
		\section{Introduction}

\begin{figure}[H]
		\centering
		\includegraphics[width=0.22\textwidth]{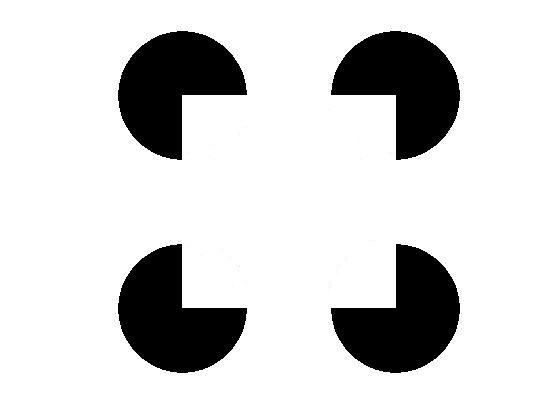}	
			\includegraphics[width=0.22\textwidth]{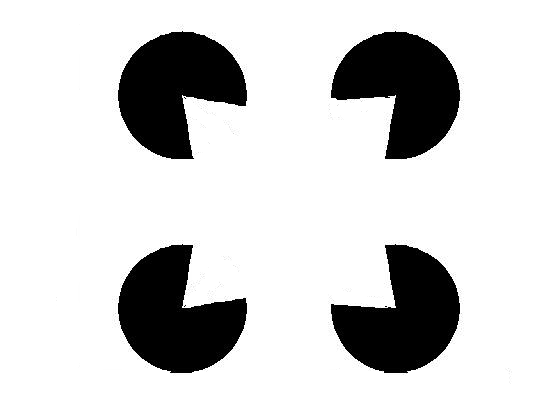}
		\includegraphics[width=0.22\textwidth]{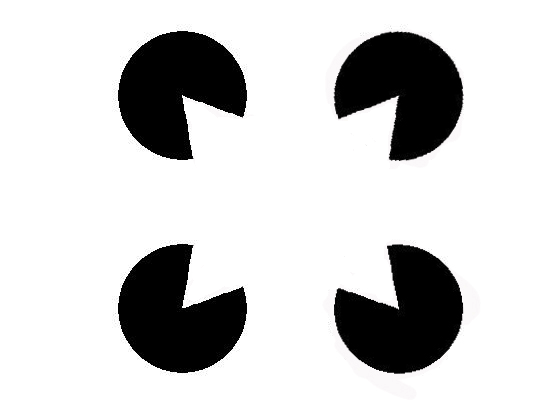}
		\includegraphics[width=0.22\textwidth]{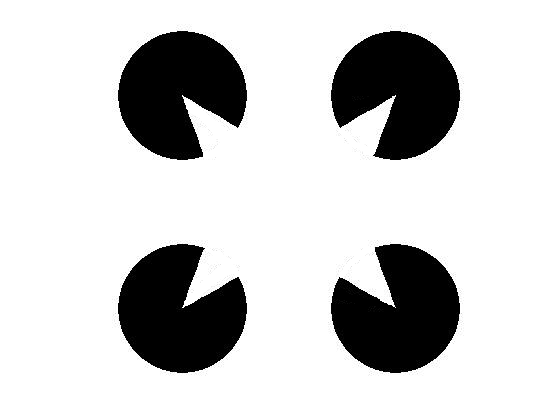}
		\caption[Experiment]{ Deformation of visual stimulus, represented by  squares with different angles between the inducers: the angle regularly decreases and we perceive regular deformations of the subjective 
		Kanizsa square up to a certain value of curvature, when the square suddenly disappears and the 
		inducers are perceived.} 
			\label{quad}
	\end{figure}

	Gestalt laws have been proposed to explain several phenomena of visual perception, such as grouping and figure-ground segmentation  (\citep{wertheimer1938laws,kohler1929gestalt,koflka1935principles} and for a recent review we quote: \citep{wagemans2012century}). In particular, in order to individuate perceptual units, gestalt theory has introduced local and global laws. Among the local laws we recall the principle of proximity, similarity and good continuation. Particularly the local law of good continuation plays a central role in perceptual grouping (see Figure \ref{quad}, left). 
	
	Regarding global laws, in the construction of percepts the feature of  saliency 
	is crucial and at the same time it escapes to easy quantitative modelling. In the Berliner Gestaltheory the concept of saliency denotes the relevance of a form with respect of a contextual frame, the power of an object to be present in the visual field. 
The role of saliency is pivotal also in figure-ground articulation. 	Due to  the perceptual grouping process 
	 the scenes are perceived as constituted by a finite number of figures and the saliency assigns a discrete value to each of them. In particular the most salient configuration pops up from the ground and becomes a figure \citep{merleau1996phenomenology}.
	Note that in case of continuous  deformation of the visual stimulus, the salient figures can change abruptly from one percept to a different one \citep{merleau1996phenomenology}.  This happens for example in Figure \ref{quad} where a regular deformation is applied to the Kanizsa square: 
	we progressively perceive a more curved  square, 
until it suddenly disappears
and the 4 inducers are perceived as stand alone (see for example \citep{lee2001,pillow2002,petitot2008}). 

	
\begin{figure}[htbp]
		\centering 
		\begin{subfigure}[b]{1.2 in} \centering
			\includegraphics[width=1.1\textwidth]{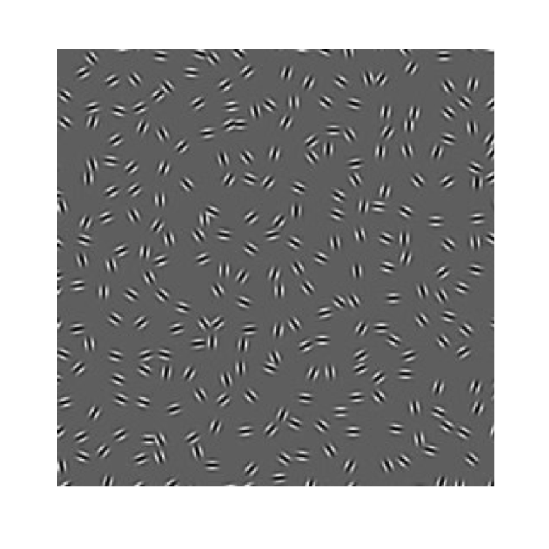}
			\caption{}
			\label{fig:gross}
		\end{subfigure}
		\begin{subfigure}[b]{1.2 in} \centering
			\includegraphics[width=1.1\textwidth]{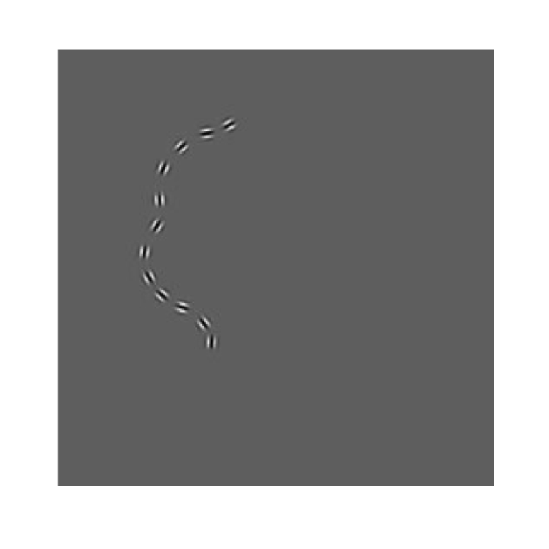}
			\caption{}
			\label{fig:field}
		\end{subfigure}
		\begin{subfigure}[b]{1.2 in} \centering
			\includegraphics[width=1.1\textwidth]{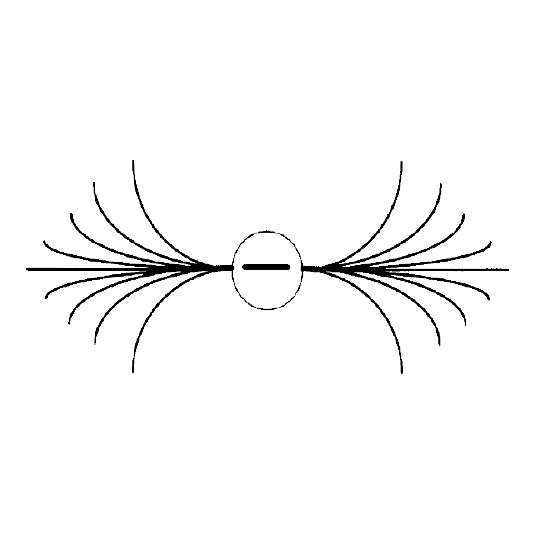}
			\caption{}
			\label{fig:WJ}
		\end{subfigure}
		\caption[Experiment]{The stimulus proposed by Field, Hayes and Hess \citep{field1993contour} (a) and the perceptual unit present in it (b). In (c) the field lines of the association field, that represents the elements in the path which can be associated to the central point \citep{field1993contour}.} \label{FHH_ex1}
	\end{figure}

	A number of results have been provided in order to refine the principles of psychology of form and assess neural correlates of the good continuation law.
	In particular, Grossberg and Mingolla in \citep{grossberg1985neural} introduced  a ``cooperation field'' 
to model illusory contour formation.
	Similar fields of association and perceptual grouping have been produced by Parent and Zucker in \citep{parent1989trace}. 
	In this contest, in the 1990s Kellman and Shipley provided a theory of object perception that specifically adressed perception of partially occluded objects and illusory contours 
	\citep{kellman1991theory,shipley1992perception,shipley1994spatiotemporal}. Heitger and von der Heydt \citep{von1993perception} provided a theory of figural completion 
	which can be applied to  both illusory contour figures (as the Kanizsa triangle) and real images.
	In the same years Field, Hayes and Hess \citep{field1993contour} introduced  through psychophysical experiments the notion of \textit{association fields}, describing the Gestalt principle of good continuation. 
	They studied how the perceptual unit visualized in  Figure \ref{FHH_ex1} (b) pops up from
 a stimulus of Gabor patches (see Figure \ref{FHH_ex1} (a)). Through a series of similar experiments, they constructed an association field, that defines the pattern of position-orientation elements of stimuli that can be associated to the same perceptual unit
	(see Figure \ref{FHH_ex1} (c)).

	Starting from the classical results of Hubel and Wiesel \citep{hubel1977ferrier} it has been possible to justify on neurophysiological bases these perceptual phenomena.  The results of \citep{bosking1997orientation} and 
	\citep{fregnac1999activity}
	confirmed that neurons sensitive to similar orientation are preferentially connected.  This suggests that the rules of proximity and good continuation are implemented in the horizontal connectivity of low level visual cortices. 
A stochastic model which takes into account the structure of the cortex, with position an orientation features, was proposed by Mumford \citep{mumford1994elastica}, and further exploited by Williams and Jacobs in  \citep{williams1997stochastic} and  \citep{august2000curve}. 
They  modelled the analogous of the association fields with 
	Fokker-Planck equations, taking into account different geometric features, such as orientation or curvature. Petitot and Tondut \citep{petitot1999vers} introduced a model of the functional architecture of V1, compatible with the association field. 
	 Citti and Sarti in  \citep{citti2006cortical} proposed the model of the functional architecture as a Lie groups, showing the relation between geometric integral curves,  association fields, and cortical properties. This method has been implemented in \citep{sanguinetti2008image} and \citep{boscain2012anthropomorphic}. Exact solution of the Fokker-Planck equation has been provided by  Duits and van Almsick \citep{duits2008explicit}, and their results have been applied by Duits and Franken \citep{duits2009line} to image processing.
	
	\smallskip
	
	The described local laws are insufficient to explain the constitution of a percept, since a perceived form is characterized by a global consistency. Different authors qualitatively defined this consistency as pregnancy or global saliency	{\citep{merleau1996phenomenology}}, but only a few quantitative models have been proposed \citep{koch1985}. In particular spectral approach for image processing were proposed by \citep{perona1998factorization,shi2000normalized,weiss1999segmentation,coifman2006diffusion}. 
	In \citep{sarti2015constitution} it is shown how this spectral mechanism is implemented in the neural morphodynamics, in terms of symmetry breaking of mean field neural equations. In that sense, \citep{sarti2015constitution} can be considered as an extension of \citep{bressloff2002geometric}.

	In this paper we further develop the approach introduced in \citep{sarti2015constitution} and describe an algorithm for the individuation of perceptual units, using both local and global constraints: local constraints are modelled by suitable connectivity kernels, which represent neural connections, and the global percepts are computed by means of spectral analysis.
		 The model is described in the geometric setting of a Lie group equipped with a Sub-Riemannian metric introduced in \citep{petitot1999vers,citti2006cortical,sarti2008symplectic}.  
		Despite the apparent mathematical difficulty, it helps to clarify in a rigorous way the gestalt law of good continuation. 
		
Here we introduce various substantial differences from the techniques in literature. 
	While studying the local properties of the model, we focus on the properties of the 
	connectivity kernels.  The Fokker Planck and the Laplacian kernel in the motion group 
	are already largely used for the description of the connectivity, since they qualitatively 
	fit the experimental data \citep{sarti2015constitution}. Here we perform a quantitative fitting between the computed kernels and the experimental ones, in order to validate the model. Moreover we propose to use 
	also the Subelliptic Laplacian kernel, in order to 
	account for the variability of connectivity patterns.   
Secondly we accomplish grouping with a spectral analysis inspired from the work of \citep{sarti2015constitution}, who proved
 the neurophysiological plausibility of this process. 
In the experiments we manipulate the stimuli 
to demonstrate the relation between the pop up of the figure and the eigenvalue analysis. We will 
 analyze in particular the swap between one solution and the other while smoothly changing the stimulus 
in many grouping experiments. 
Finally we enrich the model, exploiting the role of the 
polarity feature, which allows to work with two competing kernels.

The plan of the paper is the following. The Section 2 is divided in two parts, in the first we describe local constraints and in the second the global ones. We will first recall the neurogeometry of the visual cortex and see how the cortical connectivity is represented by the fundamental solution of Fokker Planck, Sub-Riemannian Laplacian and isotropic Laplacian equations. We propose a method for the individuation of perceptual units, first recalling the notions of spectral analysis of connectivity matrices, obtained by the connectivity kernels. We will see how eigenvectors of this matrix represent perceptual units in the image. 
	In Section 3 we present numerical approximations of the kernels and we will compare kernels with neurophysiological data of horizontal connectivity  \citep{angelucci2002circuits,bosking1997orientation}. We also perform a quantitative validation of the kernel considering the experiment of \citep{gilbert}, showing the link between the connectivity kernel and cell's response.
	Finally in Section 4 we present the results of
	 simulations using the implemented connectivity kernels. We will identify perceptual units in different Kanizsa figures, highlighting the role of polarity, discussing and comparing the behavior of the different kernels.

	\section{The mathematical model}

	In this section we identify a possible neural basis of local Gestalt laws in the functional architecture of the primary visual cortex, that is the first cortical structure that underlies the processing of the visual stimulus. 
We do not claim here that the process of grouping has to be attributed exclusively to V1, 
since several cortical areas are involved in segmentation of a figure. However 
neural evidence ensures that it 
		takes place already in V1 (see \citep{lee2001, pillow2002}). Hence we focus on this 
		area where the 
		first elaboration is made and it 
		is particular important for all the geometrical aspects of the process.
	
	\subsection{Local constraints - The neurogeometry of V1}

	In the 70s Hubel and Wiesel  discovered that this cortical area is organized 
	in the so called hypercolumnar structure  (see \citep{hubel1962receptive,hubel1977ferrier}). This means that 
	for each retinal point $(x, y)$ there is an entire set of cells each one sensitive to a 
	specific orientation $\theta$  of the stimulus. 
	
	The first geometrical models of this structure are due to Hoffman \citep{hoffman1989visual}, Koenderink \citep{koenderink1987representation}, Williams and Jacobs \citep{williams1997stochastic} and Zucker \citep{zucker2006differential}. 
	They described the cortical space as a fiber bundle, where the retinal plane $(x,y)$ is the basis, while the fiber concides with the hypercolumnar variable $\theta$. 
	More recently Petitot and Tondut \citep{petitot1999vers}, Citti, Sarti \citep{citti2006cortical} and Sarti, Citti, Petitot \citep{sarti2008symplectic}, proposed to  describe this structure as a Lie group with a Sub-Riemannian metric (see also the results of \citep{duits2009line}). This expresses the fact that  each filter can be recovered from a fixed one by translation of the point $(x,y)$ and rotation of an angle $\theta$. In particular the visual cortex  can be described as the subset of points of  $\mathbb{R}^2\times\mathit{S}^1$.	Every simple cell is characterized by its receptive field, 
	classically defined as the domain of the retina to which the neuron is sensitive. The shape of the 
	response of the cell in presence of a visual input is called receptive profile (RP) and can be reconstructed by electrophysiological recordings \citep{ringach2002spatial}. In particular simple cells of V1 
	are sensitive to orientation and are strongly oriented. Hence their RPs are interpreted as Gabor patches \citep{daugman1985uncertainty,jones1987evaluation}. 
	Precisely they are constituted by two coupled families of cells: an even and an odd-symmetric one.
	
	Via the retinotopy, the retinal plane can be identified with the 2-dimensional plane $\mathbb{R}^2$. A visual stimulus at the retinal point $(x,y)$ activates the whole hypercolumnar structure over that point. All cells fire, but the cell with the same orientation of the stimulus is maximally activated, giving rise to orientation selectivity.	

Formally curves and edges are lifted to new cortical curves, identified by the variables $(x,y,\theta)$, where $\theta$ is the direction of the boundary at the point $(x,y)$. In \citep{citti2006cortical} it has been shown that these curves are always tangent to the planes generated by the vector fields.
These curves have been modelled by  \citep{citti2006cortical} as integral curves of 
suitable vector fields in  the $SE(2)$ cortical structure. Precisely, the vector fields they considered
are: 
\numberwithin{equation}{section}
\begin{equation}
	\vec X_1 = (\cos\theta,\sin\theta,0),  \mbox{\quad}     \vec X_2 = (0,0,1).
	\label{vector}
\end{equation}
All lifted curves are integral curves of these two vector fields
such that a curve in the cortical space is: 
\begin{equation}
	c '(s)=(k_1(s)\vec X_1 + k_2(s)\vec X_2)(c(s)),\mbox{\quad} c(0)=0. 
	\label{eq}
\end{equation}
It has been noted in \citep{citti2006cortical} that these curves, projected on the 
2D cortical plane are a good model of the association fields. 

\subsection{A model of cortical connectivity}

From the neurophysiological point of view, there is experimental evidence of the existence of connectivity between simple cells belonging to different hypercolumns. It is the so called long range horizontal connectivity. Combining optical imaging of intrinsic signals with small injections of biocytin in the cortex, Bosking et al. in  \citep{bosking1997orientation} led to clarify properties of horizontal connections on V1 of the tree shrew. 
The propagation of the tracer is strongly directional and the direction of propagation coincides with the preferential direction of the activated cells. 
Hence connectivity can be summarized as preferentially linking neurons with co-circularly aligned receptive fields. 

The propagation along the connectivity 
can be modeled as the stochastic counter part of the deterministic curves defined in Eq.(\ref{eq}) for the description of the output of simple cells.  If we assume a deterministic component in direction  $X_1$ (which describes the long range connectivity) and stochastic component along  $X_2$ (the direction of intracolumnar connectivity), the equation can be written as follows:
\begin{equation}
	(x',y',\theta') = (\cos\theta,\sin\theta,N(0,\sigma^2)) = \vec X_1 + N(0,\sigma^2)\vec X_2
	\label{eq_stoc}
\end{equation}
where $N(0,\sigma{^2})$ is a normally distributed variable with zero mean and variance equal to $\sigma^2$. 
The probability density of this process, denoted by $v$, 
 was first used by Williams and Jacobs \citep{williams1997stochastic} to compute stochastic completion field, by August and Zucker \citep{august2000curve,august2003sketches}  to define the curve indicator random field, and more recently by R. Duits and Franken in \citep{duits2008explicit,duits2009line} to perform contour completion, de-noising and contour enhancement. 
%
The kernel obtained integrating in time the density $v_1$
\begin{equation}\Gamma_1(x,y, \theta) = \int_0^{+\infty} v_1(x,y, \theta, t) dt
\label{GammaFP} \end{equation} 
is the fundamental solution of the Fokker Planck operator
$
	FP = X_1 +\sigma^2X_{22}.
$

\begin{figure}[htbp]
	\centering 
	\begin{subfigure}[b]{1.3 in} \centering
		\includegraphics[width=1.2\textwidth]{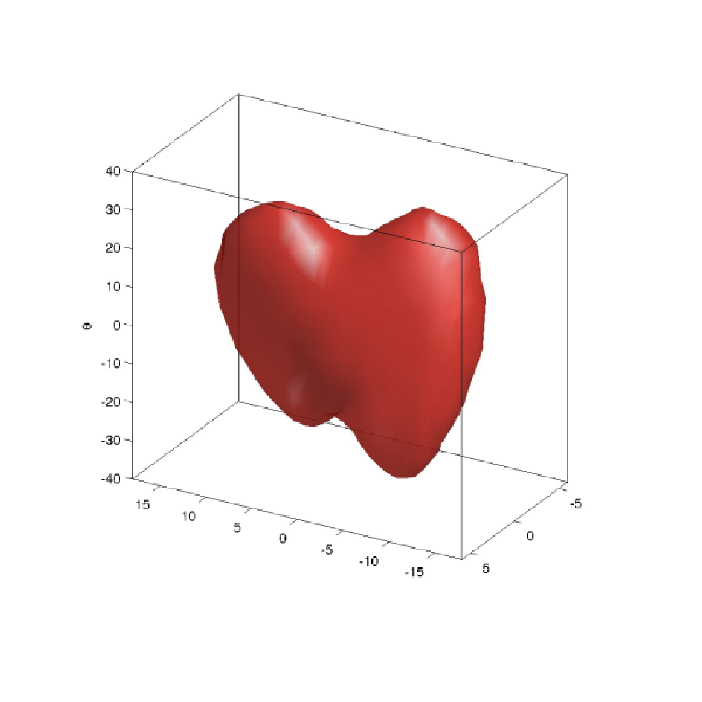}
		\caption{}
		\label{fig:gross}
	\end{subfigure}
	\begin{subfigure}[b]{1.3 in} \centering
		\includegraphics[width=1.2\textwidth]{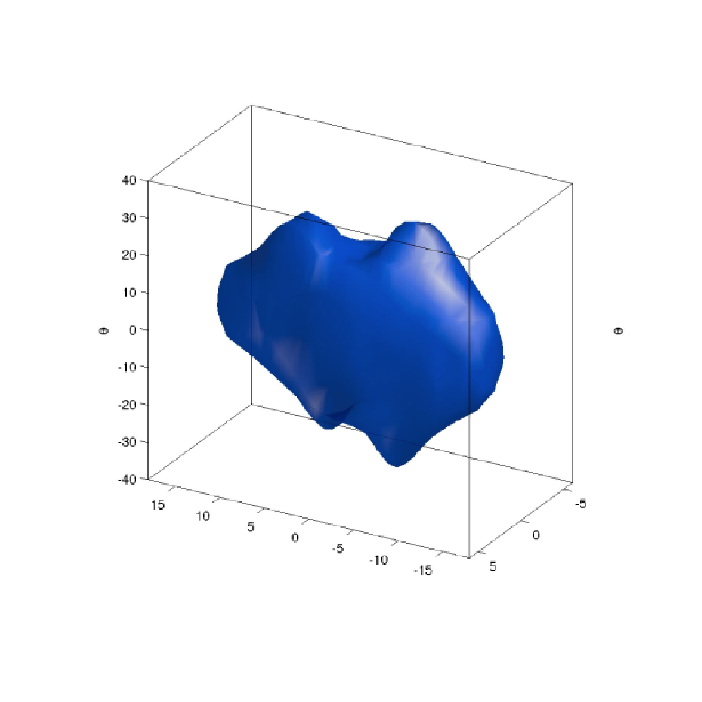}
		\caption{}
		\label{fig:field}
	\end{subfigure}
	\begin{subfigure}[b]{1.4 in} \centering
		\includegraphics[width=1.2\textwidth]{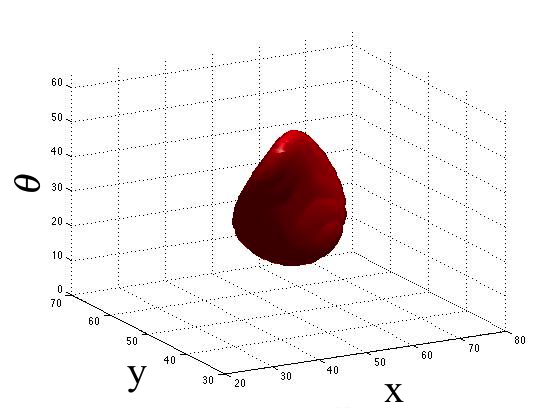}
		\caption{}
		\label{fig:vH}
	\end{subfigure}
	\begin{subfigure}[b]{1.4 in} \centering
		\includegraphics[width=1.2\textwidth]{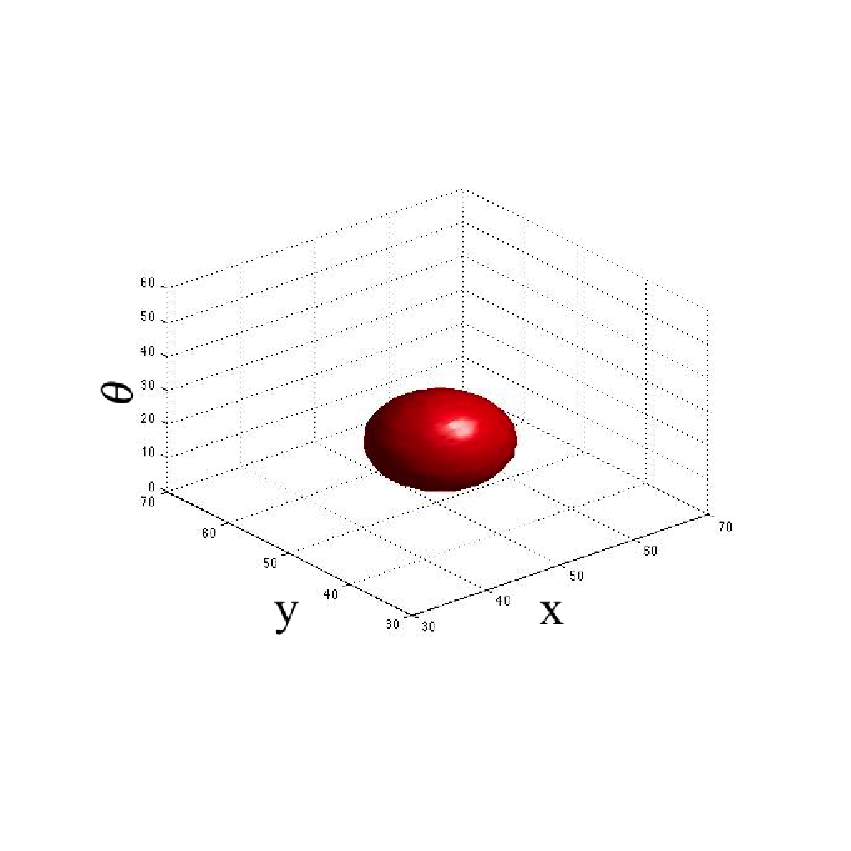}
		\caption{}
		\label{fig:WJ}
	\end{subfigure}
	\caption[Fundamental solutions]{An isosurface of the connectivity kernel $\omega_1$ obtained by symmetrization of the Fokker Planck fundamental solution Eq.(\ref{GammaFP}) (a). The distribution of co-occurrence of edges in natual images (from \citep{sanguinetti2008image}) (b). Isosurface of the connectivity kernel $\omega_2$ obtained from the fundamental solution $\Gamma_2$ of the Sub-Riemannian Laplacian equation Eq.(\ref{GammaSRL}) (c). An isosurface of the fundamental solution of the isotropic Laplacian Eq.(\ref{GammaISO}) (d).}\label{ISO}
\end{figure}

The kernel $\Gamma_1$  is strongly biased in direction $X_1$ and not symmetric; a new symmetric kernel can be obtained as following:
\begin{equation}
	\omega_1((x,y,\theta),(x',y',\theta')) = \frac{1}{2}(\Gamma_1((x,y,\theta),(x',y',\theta'))+\Gamma_1((x',y',\theta'),(x,y,\theta)).
	\label{omega}
\end{equation}
In Figure \ref{ISO} (a) it is visualized an isosurface of the simmetrized kernel $\omega_1$, showing its typical twisted butterfly shape. 
The kernel $\omega_1$ has been proposed in \citep{sanguinetti2008image} as a model of the  statistical distribution
 of edge co-occurrence in natural images, as described in (Sanguinetti et al., 2008). The similarity between the two is proved both at a qualitative and at a quantitative level (see (Sanguinetti et al., 2008)) (see also Figure \ref{ISO} (a) and (b)).

If we assume that intracolumnar and long range connections have comparable strength, the stochastic 
equation Eq.(\ref{eq_stoc}) reduces to:
\begin{equation}
	(x',y',\theta') = N(0,\sigma_1^2)\vec X_1 + N(0,\sigma_2^2)\vec X_2 
	\label{eq_SRL}
\end{equation}
where $N(0,\sigma_i{^2})$ are normally distributed variables with zero mean and variance  equal to $\sigma_i^2$. In this case the speed of propagation in directions $X_1$ and $X_2$ is comparable. The associated probability density is the fundamental solution of the Sub-Riemannian Heat equation \citep{Jerison1987}. The integral in time of this probability density 
\begin{equation}\Gamma_2(x,y, \theta) = \int_0^{+\infty} v_2(x,y, \theta, t) dt\label{GammaSRL}\end{equation}
is the fundamental solution of the Sub-Riemannian Laplacian (SRL): 
$
	SRL = \sigma_1^2X_{11} + \sigma_2^2X_{22}.
	\label{eq_srlap}
$

It is a symmetric kernel, so that we do not need to symmetrize it and we use it as a model of the connectivity kernel:
\begin{equation}
	\omega_2((x,y,\theta),(x',y',\theta')) = \Gamma_2((x,y,\theta),(x',y',\theta')).
\end{equation}
In Figure \ref{ISO} (c) it is shown an isosurface of the connectivity kernel $\omega_2$.


We will see in Section 3.2 that a combination of Fokker-Planck  and Sub-Riemannian Laplacian 
fits the connectivity map measured by Bosking in \citep{bosking1997orientation}, where the Fokker-Planck fundamental solution represents well the long distances of the trajectory, while the Sub-Riemannian Laplacian the short ones. 
Combination of different Fokker-Planck fundamental solutions can also be used to model the functional architecture of primates experimentally  measured by Angelucci in \citep{angelucci2002circuits}.

While validating the model, we will 
see that a standard Riemannian kernel does not provide the same 
accurate results. In order to show this
we will introduce an isotropic version of the previous 
model which is a standard Riemannian kernel. 
To constuct it, we complete the family of vector 
fields in Eq.(\ref{vector}) with an orthonormal one:
\begin{equation}
	\vec X_3 = (-\sin(\theta), \cos(\theta), 0)
\end{equation}
choosing stochastic propagation in any direction, in such a way that equation Eq.(\ref{eq_stoc}) becomes: 
\begin{equation}
	(x',y',\theta') = N(0,\sigma^2)\vec X_1 + N(0,\rho^2)\vec X_2 +N(0,\sigma^2)\vec X_3,
	\label{eq_iso}
\end{equation}

where $N(0,\sigma_i{^2})$ are normally distributed variables with zero mean and variance  equal to $\sigma_i^2$. 
Its probability density will be denoted $v_3$ and the associated time independent kernel 
\begin{equation}\Gamma_3(x,y, \theta) = \int_0^{+\infty} v_3(x,y, \theta, t) dt\label{GammaISO}\end{equation}
will be the fundamental solution of the standard Laplacian operator:
$
	L = \sigma^2X_{11} + \rho^2X_{22} + \sigma^2X_{33} = \sigma^2(\partial_{xx}+\partial_{yy})+\rho^2\partial_{\theta\theta}.
$
One of its level sets is represented in Figure \ref{ISO} (d).

In section 3.1 we will describe a numerical technique to construct the 3 kernels described above.

\subsection{Global integration}

Since the beginning of the last century perception has been considered by gestaltist as a global process. Moreover, following Koch-Ullman and  Merleau-Ponty, visual perception is a process 
 of the visual field, that individuates figure and background at the same time \citep{koch1985, merleau1996phenomenology}. Then it continues in segmentation of the structures by succeeding differentiations.

A cortical mechanism responsible for this analysis has been outlined by  
\citep{sarti2015constitution}, starting from the classical mean field equation 
of Ermentrout and Cowan \citep{ErmentroutCowan} and Bressloff and Cowan 
\citep{bressloff2002geometric,bressloff2003functional}. 
This equation describes the evolution of the cortical activity and depends on the connectivity kernels.  
The discrete output $h$ of the simple cells, 
selects in the cortical space $(x,y,\theta)$ the set of active cells 
and the cortical connectivity, restricted on this set, 
defines a neural affinity matrix. 
The eigenvectors of this matrix 
describe the stationary states of the mean field equation 
hence the emergent perceptual units. 
The system will tend to the eigenvector associated to the highest eigenvalue, which corresponds to the most important object in that scene. 
Mathematically the approach is strongly linked to spectral analysis techniques for locality-preserving embeddings of large data sets \citep{coifman2006diffusion,belkin2003laplacian,roweis2000nonlinear}, for data segregation and partitioning \citep{perona1998factorization,meila2001random,shi2000normalized}, 
{grouping process in real images \citep{weiss1999segmentation}.}

\subsection{The cortical activity equation}

We have seen that in presence of a visual stimulus 
cells aligned to its boundary give the maximal response. 
We will assume that a discrete number of cells $N$ are maximally activated and we will denote them $(x_i,y_i,\theta_i)$ for $i = 1,...,N$. 
In Figure \ref{firstexp} (b) we show as an example the cells responding to a Kanizsa figure, 
represented with their Gabor-like receptive profiles. 
Following \citep{sarti2015constitution} the cortical connectivity is restricted to this discrete set 
and reduces to a matrix  $A$: 
\begin{equation}
	A_{i,j} = \omega((x_i,y_i,\theta_i), (x_j,y_j,\theta_j)).
	\label{Maff_eq}
\end{equation}

In this discrete setting the mean field equation 
for the cortical activity reduces to:
\begin{equation} \label{eqdiscrete}
\frac{du}{dt}=- \lambda u(i)+
 s\Big(\sum_{j=1}^N A(i,j)u (j)\Big)  
\end{equation}
where $s$ is a sigmoidal function and $\lambda$ is a physiological parameter. 
The solution tends to its stationary states, which 
are the eigenvectors of the associated linearized equation: 
\begin{equation}
\sum_{j=1}^N 	A_{i,j}u_j = \frac{\lambda}{s'(0)} u_i
\end{equation}

Hence these are the emergent states of the cortical activity, that 
individuate the coherent perceptual unit in the scene and allow to segment it. 
This is why we will assign to the eigenvalues of the affinity matrix the meaning of a saliency index of the objects. 
Since we have defined three different kernels 
different affinity matrices will be defined. However all kernels are real and symmetric, so that the matrix $A$ is a real symmetric matrix $A_{i,j}$ = $A_{j,i}$. Their eigenvalues are real and the highest eigenvalue is defined. 
The associated principal eigenvectors emerge as symmetry breaking of the stationary solutions of mean fields equations and they pop up abruptly as emergent solutions. 
The first eigenvalue will correspond to the most salient object in the image.

\subsection{Individuation of perceptual units}

Since the three different kernels assign different role to different direction of connectivity, the different affinity matrices and their spectrum will reflect these different behavior. Consequently the resulting data set partitioning will be stronger in the straight direction using the Fokker Planck $\omega_1$ kernel, or will allow rotation using the $\omega_2$ kernel (see also \citep{cocci2015cortical} for a deeper analysis). Using the kernel $\omega_3$ we expect an equal grouping capability in the collinear direction and in the ladder direction. 

\begin{figure}[H]
	\centering
	\includegraphics[width=0.35\textwidth]{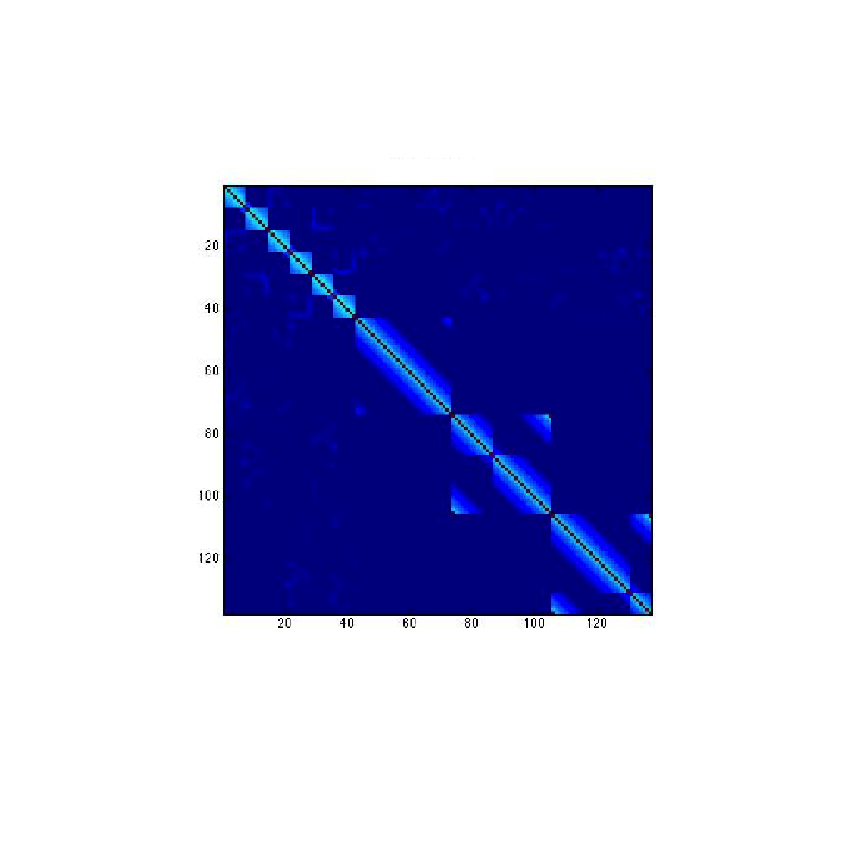}
	\includegraphics[width=0.35\textwidth]{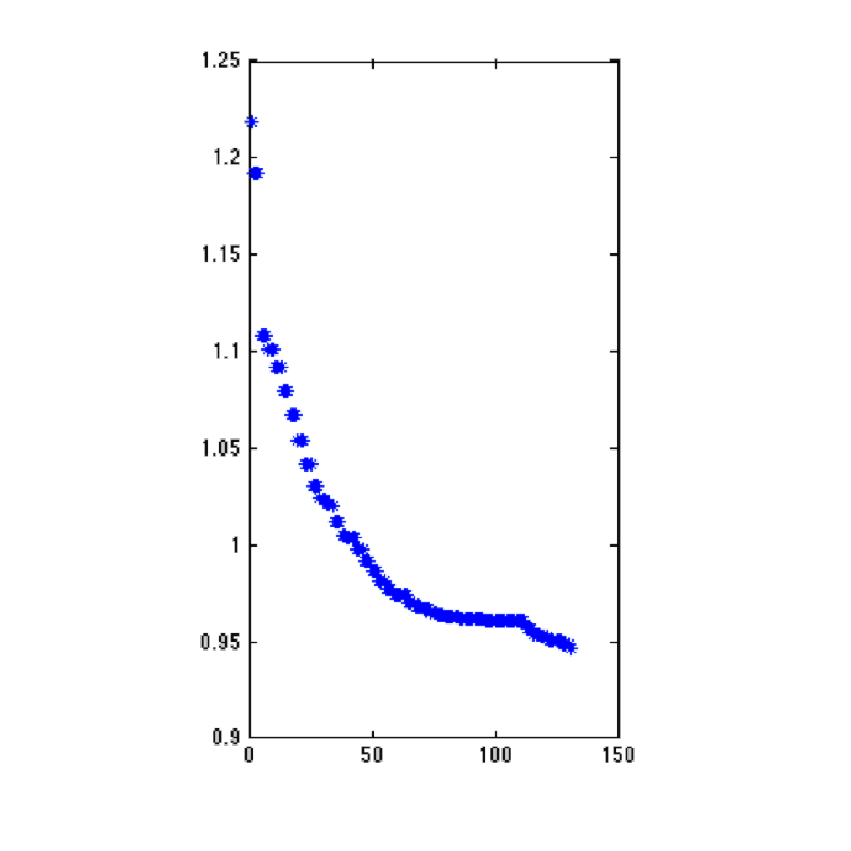}
	\caption{On the left it is visualized the affinity matrix that contains informations about the affinity of an active patch with respect to all the others. On the right the set of its sorted eigenvalues.}
	\label{Maff}
\end{figure}

In Figure \ref{Maff} we visualize the affinity matrix of the image presented in Figure \ref{firstexp} (a).  It is a square matrix with dimensions NxN, where $N$ is exactly the number of active patches. It represents the affinity of each patch with respect to all the others.
The structure of the affinity matrix is composed by blocks and the principal ones corresponds to coherent objects. On the right we visualize the complete set of eigenvalues in a graph (eigenvalue number, eigenvalues). 
Let us explicitly note that the first eigenvector will have the meaning of emergent 
perceptual unit. The other eigenvectors do not 
describe an ordered sequence of figures with different rank. However, 
their presence is important, above all when two eigenvalue 
have similar values. In this case, small deformation of the stimulus 
can induce a change in the order of the eigenvalues and produce a sudden emergence 
of the correspondent eigenvector with an abroupt change in the perceived image. 
%
%

This is in good agreement with the perceptual characteristics of salient figures of temporal and spatial discontinuity, since they pop up abruptly from the background, while the background is perceived as indifferentiated \citep{merleau1996phenomenology}. Spectral approaches give reason to the discontinuous character of figure-ground articulation better than continuous models, who instead introduce a graduality in the perception of figure and background \citep{lorenceau2001form}.
\\
To find the remainig objects in the image, the process is then repeated on the vector space orthogonal to $p$, the second and the following eigenvectors can be found, until the associated eigenvalue is sufficiently small.
In this way only $n$ eigenvectors are selected, with  $n<N$, this procedure reduces the dimensionality of the description. This procedure neurally reinterprets the 
process introduced by Perona and Freeman in \citep{perona1998factorization}. 

%
%

\section{Quantitative kernel validations}
\subsection{Numerical approximations of the kernels}

In this section we numerically approximate the connectivity kernels $\omega_i$, defined in Section 2.\\ 
We obtain the discrete fundamental solution $\Gamma_1$ of Eq.(\ref{GammaFP}) by developing random paths from the numerical solution of the system (\ref{eq_stoc}), that can be approximated by:
\begin{equation}
\begin{cases} 
x_{s+\Delta{s}} - x_s = \Delta{s} \cos(\theta) \\ 
y_{s+\Delta{s}} - y_s = \Delta{s} \sin(\theta) , \mbox{\quad} s\in{0,...,H} \\
\theta_{s+\Delta{s}} - \theta_s = \Delta{s} N(\sigma,0)
\end{cases}
\label{system_FP}
\end{equation}

where $H$ is the number of steps of the random path and $N(\sigma,0)$ is a generator of numbers taken from a normal distribution with mean 0 and variance $\sigma$.
In that way, the kernel is numerically estimated with Markov Chain Monte Carlo methods (MCMC) \citep{robert2013monte}.   Various realizations  $n$ of the stochastic path will be given solving this finite difference equation  $n$ times; the estimated kernel is obtained averaging their passages over discrete volume elements, as described in detail in \citep{higham2001algorithmic,sarti2015constitution}. 
Proceeding with the same methodology the numerical evaluation of fundamental solution $\Gamma_2$ of the hypoelliptic Laplacian (Eq.(\ref{GammaSRL})) is obtained and the system (\ref{eq_SRL}) discretized:
\begin{equation}
\begin{cases} 
x_{s+\Delta{s}} - x_s = \Delta{s}R\cos(\theta) \\ 
y_{s+\Delta{s}} - y_s = \Delta{s}R\sin(\theta) , \mbox{\quad}    s\in{0,...,H} \\
\theta_{s+\Delta{s}} - \theta_s = \Delta{s}N(\sigma_3,0)
\end{cases}
\label{system_SRL}
\end{equation}
where $R = N(\sigma_1,0)$ and $\sigma_3$ is the variance in the $\theta$ direction. The kernel represented in Figure \ref{ISO} (c) is obtained by the numerical integration of that system and averaging as before the resulting paths.\\
Finally, the system (\ref{eq_iso}), that is a model for isotropic diffusion equation (Eq. \ref{GammaISO}), is approximated by:
\begin{equation}
\begin{cases} 
x_{s+\Delta{s}} - x_s = \Delta{s}N(\sigma,0) \\ 
y_{s+\Delta{s}} - y_s = \Delta{s}N(\sigma,0) , \mbox{\quad}    s\in{0,...,H} \\
\theta_{s+\Delta{s}} - \theta_s = \Delta{s}N(\rho,0)
\end{cases}
\label{system_L}
\end{equation}
where $\sigma$, $\rho$ are the variances in the $x$, $y$, $\theta$ directions. In order to obtain the approximation of the kernel $\omega_3$, visualized in Figure \ref{ISO} (d), the system is integrated with the same technique used before. \\
These kernels will be used to construct the affinity matrices in Eq.(\ref{Maff_eq}).

\subsection{Stochastic paths and cortical connectivity}

We  will now study in which extent kernels $\omega_i$, $i = 1,2$ are models of connectivity. 
The kernel $\omega_3$ is used for comparison and to show that 
an uniform Euclidean kernel does not capture the anysotropic structure of the cortex. 
 Random paths that we compute through MCM are implemented in the functional architectures in terms of horizontal connectivity of a single cell. On the other hand the connectivity of an entire population of cells corresponds to the set of all single cells connectivities, then to the Fokker Planck fundamental solution.

A first qualitative comparison between the kernels $\omega_1$, $\omega_2$ and the 
connectivity pattern has been provided in \citep{sarti2015constitution}. Here we follow the same framework, but we propose a 
more accurate, quantitative comparison. 

\begin{figure}[H]
	\centering 
	\begin{subfigure}[b]{1.4 in} \centering
		\includegraphics[width=1 \textwidth]{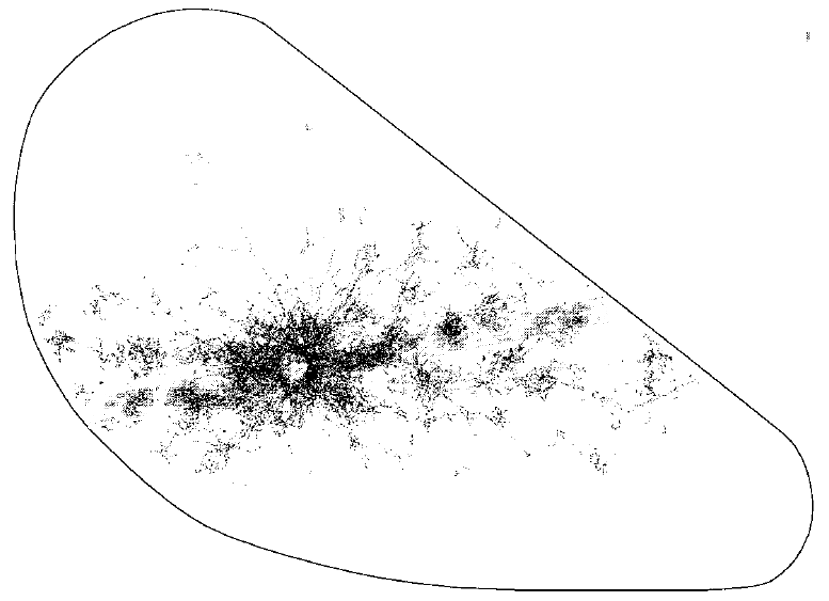}
		\caption{}
		\label{fig:B1}
	\end{subfigure}
	\begin{subfigure}[b]{1.4 in} \centering
		\includegraphics[width=1 \textwidth]{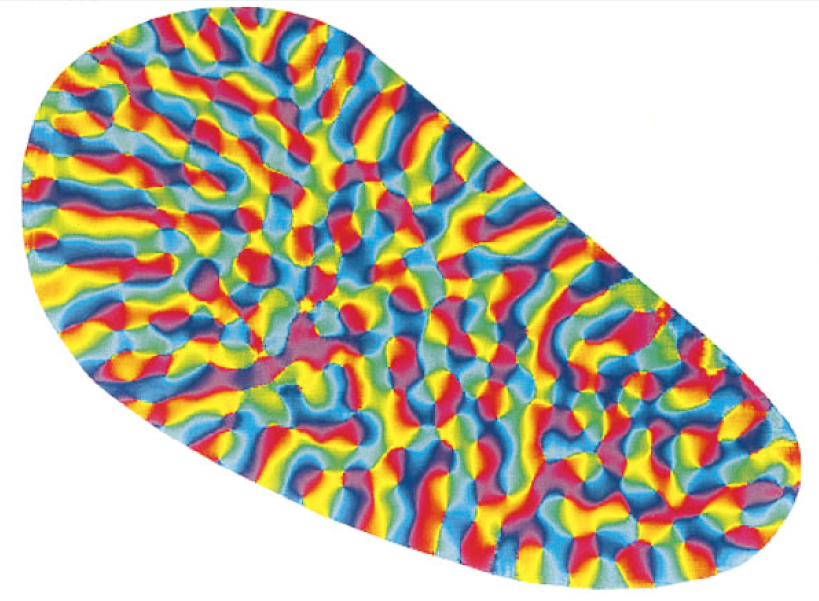}
		\caption{}
		\label{fig:B2}
	\end{subfigure}
	\begin{subfigure}[b]{1.4 in} \centering
		\includegraphics[width=1\textwidth]{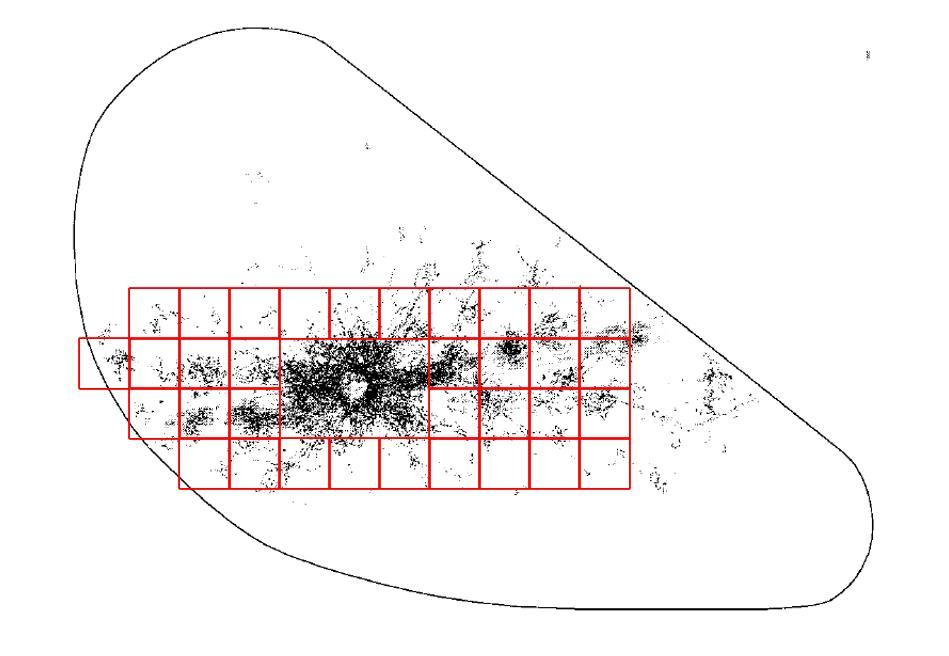}
		\caption{}
		\label{fig:B3}
	\end{subfigure}
	\begin{subfigure}[b]{1.4 in} \centering
		\includegraphics[width=1.25\textwidth]{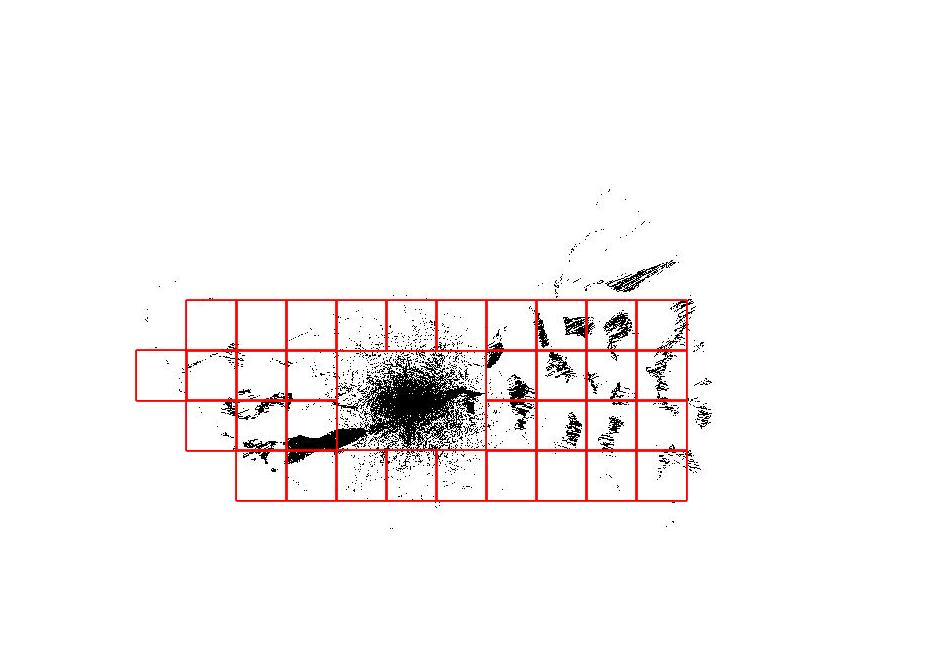}
		\caption{}
		\label{fig:B4}
	\end{subfigure}
	\\
		\centering 
	\begin{subfigure}[b]{1.4 in} \centering
		\includegraphics[height=1\textwidth]{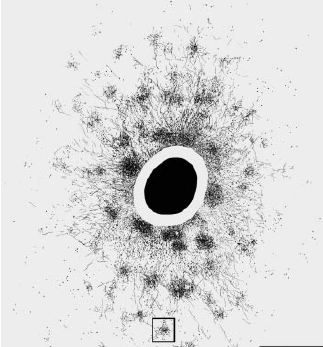}
		\caption{}
		\label{fig:B1}
	\end{subfigure}
	\begin{subfigure}[b]{1.4 in} \centering
		\includegraphics[height=1\textwidth]{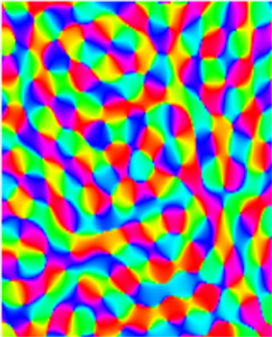}
		\caption{}
		\label{fig:B2}
	\end{subfigure}
	\begin{subfigure}[b]{1.4 in} \centering
		\includegraphics[height=1  \textwidth]{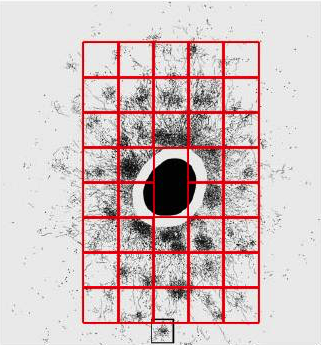}
		\caption{}
		\label{fig:B3}
	\end{subfigure}
	\begin{subfigure}[b]{1.4 in} \centering
		\includegraphics[width=1.15 \textwidth]{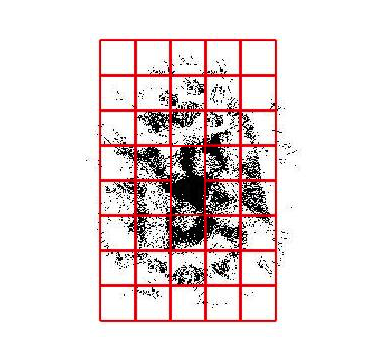}
			\caption{}
		\label{fig:B4}
	\end{subfigure}
	\\
		\centering 
	\begin{subfigure}[b]{1.4 in} \centering
		\includegraphics[height=1.05\textwidth]{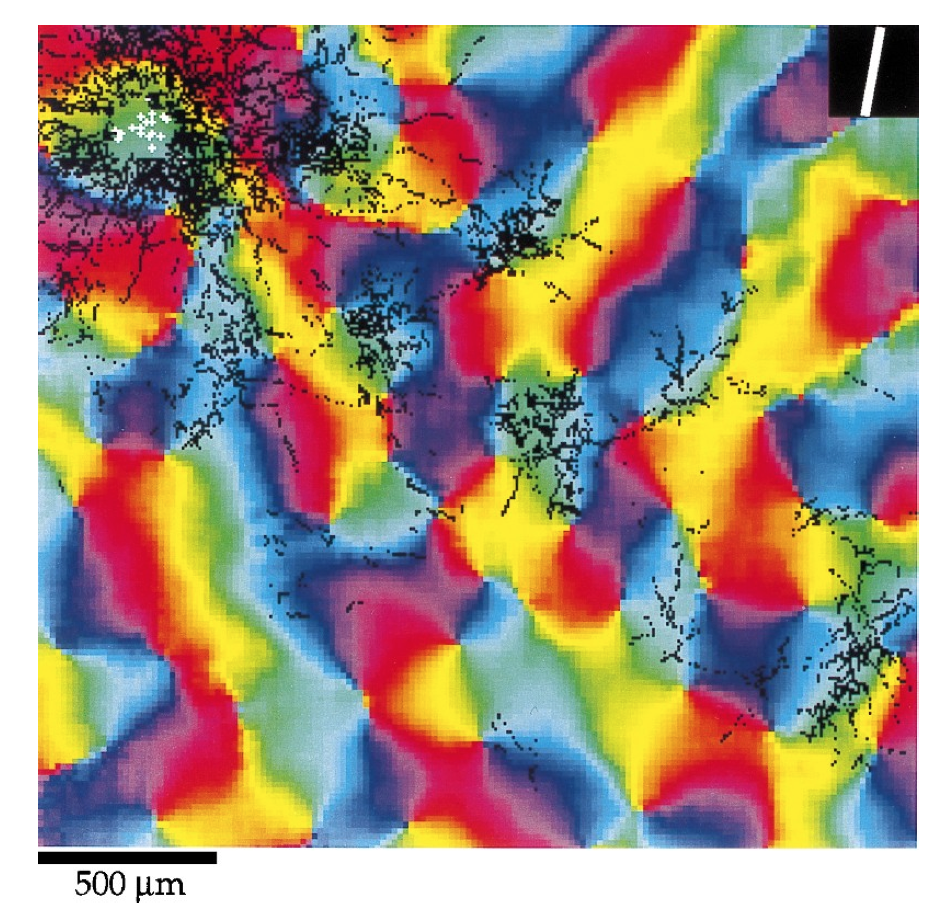}
		\caption{}
		\label{fig:B1}
	\end{subfigure}
	\begin{subfigure}[b]{1.4 in} \centering
		\includegraphics[height=1.1\textwidth]{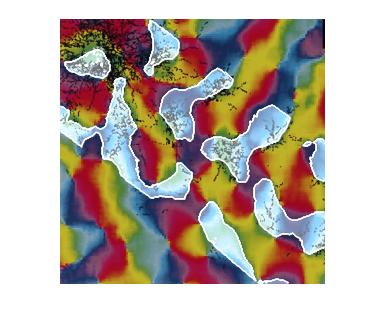}
		\caption{}
		\label{fig:B2}
	\end{subfigure}
	\caption[Max]{ The connectivity map measured by Bosking \citep{bosking1997orientation} (a) and Angelucci \citep{angelucci2002circuits} (e), the pinwheel structure used for the estimate ((b) and (f)), the tracer partitioned according to rectangles with sides equal to the distance between pinwheels ((c) and (g)) and the best fit results ((d) and (h)). In (i) the tracer superimposed to the piwheel structure found by \citep{bosking1997orientation} and in (j) the isocontours obtained from a combination of Fokker Planck.} \label{maps}

\end{figure}

It is well known that the 3D cortical structure is implemented in 
the 2D cortical layer as a pinwheel structure, which codes for position and orientations (see Figure \ref{maps} (b)).
The pinwheel structure has a large variability from one subject to one other, 
but within each species common statistical properties have been 
obtained.
Cortico-cortical connectivity has been measured by Bosking in \citep{bosking1997orientation} by 
injecting a tracer in a simple cell and 
recording the trajectory of the tracer. In Figure \ref{maps} (a) the propagation 
through the lateral connections is represented by black points. 
Bosking found a large variability of injections, which however have 
common stochastic properties as the 
direction of propagation, a patchy structure with small blobs 
at approximately fixed distance and the decay of the density of tracer along the injection site. 
 
We model each injection with stochastic paths solutions of Eq. (\ref{eq_stoc}). 
Then we evaluate the stochastic paths on the pinwheel structure. 

Due to the stochastic nature of the problem, we do not compare 
pointwise the image of the tracer and the stochastic paths 
but we average them on the pinwheels. 
We partition both the images of the tracer and of the 
stochastic paths in $M$ regions corresponding to the pinwheels: 
\begin{equation}\label{regioni}I= \cup_i R_i\end{equation}
and for every $R_i$ we compute the density of tracer $DT_i$ and the density of the 
stochastic paths $DP_i$. 
The two vectors 
$DT_i$ and $DP_i$ are then rescaled in such a way to have unitary $L^2$-norm and 
the mean square error is computed:
\begin{equation}\label{err}E = \sqrt{\frac{1}{M} \sum\limits_{i=1}^M \Big({DP_i-DT_i}\Big)^2} \end{equation}
The free parameters of the model are the value of the standard deviation, the number of paths, the number of steps, appearing in Eq.(\ref{eq_stoc}) and in the system (\ref{system_FP}). 
The best fit between the experimental and simulated distributions
has been accomplished by minimizing the mean square error by varying these parameters. 

\bigskip

Due to the different role of the directions $X_1$ and $X_2$ in the definition of these kernels, 
the Sub-Riemannian Laplacian paths and the Fokker Planck paths 
have different structure. 

The Subriemannian Laplacian allows diffusion in direction $X_2$, 
favors the changement of the angle and it 
can be used to describe short range connectivity as described in Section 4.4. Hence it is responsible 
for the central blob, in 
a neighborhood of the injection points (see Figure \ref{maps} (c)). 
The Fokker Planck kernel produces an elogated, patchy structure and seems responsible 
for the long range connection (see Figure \ref{maps} (d)). 
 We apply our quantitative fit only to the long range connectivity, 
hence discarding the tracer in a neighborhood of the injection. For this reason the Sub-Riemannian Laplacian is not involved in the validation of the model.

 The method is first applied to fit the image of the tracer taken by Bosking 
\citep{bosking1997orientation} (see Figure \ref{maps} (a)). 
All the kernels are evaluated on the pinwheels 
provided in the same paper (see figure \ref{maps} (b)), to obtain a patchy structure. 
In  order to apply the formula \eqref{regioni}, 
we cover both the image of the tracer and the Fokker Planck with 
a regular distribution of rectangles, with edges equal 
to the mean distances between pinwheels (see Figure \ref{maps} (c),(d)) (clearly we do not cover the central zone, where we can not fit the Fokker Planck kernel). 
The resulting 
error value is $E<8\%$, 
showing that the model accurately represents the
experimental distribution. 
 
 A similar procedure has been applied to the image of the tracer provided in 
\citep{angelucci2002circuits} (see Figure \ref{maps} (e)). 
The result of Angelucci is obtained with various injections in a neighborhood of a 
pinwheel, so that all orientations are present, and the tracer propagates in all directions.
In this case we do not have natural pinwheels, hence we use artificial pinwheels, 
obtained with the algorithm presented in \citep{barbieri2012} (see Figure \ref{maps} (f)), with the constraint that the mean distance between the artificial pinwheels is equal to the mean distance between the blobs produced by the tracer. 
Here we consider Fokker Planck paths with all directions, to obtain the 
apparent isotropic diffusion. 
Also in this case we cover with rectangles and perform a best fit and the minimum error value is $E< 8\%$, 
(see figure \ref{maps} (g), (h).
 
 In his paper \citep{bosking1997orientation} Bosking showed a famous 
image, with the tracer superimposed to the piwheel structure (see Figure \ref{maps} (i)). In particular in this case we have the tracer and the pinwheel 
of the same animal. This allows to go below the scale of the pinwheel  
and we correctly recover the orientation with the pinwheel (see Figure \ref{maps} (j)). The estimated kernel is again a combination of Fokker Plank. As before, we focus on orientations, hence we only model the long range 
part of the image, discarding the central blob. 
The evaluation of the error is made with squared regions at a scale
smaller that the pinwheel and the error goes below $E< 9\%$.

\subsection{Perceptual facilitation and density kernels}

In order to obtain a stable and deterministic estimate of this stochastic model, we used the density kernel, which is a regular deterministic function, 
coding the main properties of the process. 
We perform here a quantitative validation of these regular kernel comparing to
an experiment of \citep{gilbert}. 

This work studies the capability of cells 
to integrate information out of the single receptive field of the cells. 
This integration process is due to the long-range
horizontal connections, hence it can be used to validate our 
model of long range connectivity. As we have recognized in the previous section 
it is the Fokker Planck kernel which can be considered as a model for 
long range connectivity, hence we use here this kernel.

	\begin{figure}[H]
		\centering 
		\includegraphics[width=0.55\textwidth]{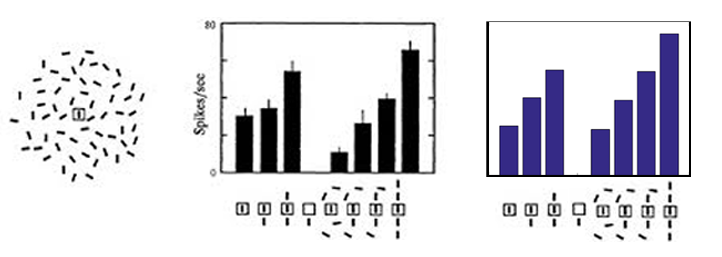}
		\caption[hist]{ On the left the experiment of \citep{gilbert}, with the stimulus composed by  randomly placed and oriented lines and the black histogram of cell's response. On the right the histogram evaluated from the probability density in response to the same distribution of lines.

		} \label{hist}	
	\end{figure}
	
	In Figure \ref{hist} (left) it is shown the results of \citep{gilbert}, where it is visualized the cell's response to randomly placed and oriented lines in a black histogram. 
	 A vertical line is present in the receptive field of a cell selective to this orientation and 
	the intensity of its response 
	is represented in the first column of the histograms. If the stimulus is surrounded by random elements aligned with 
	the first one, the cell's response increases (respectively in the second, third and the last column of the histograms). When the other random elements are not aligned with the fixed one (as in the fifth, sixth, seventh columns), the cell's response decreases because it reflects an inhibitory effect. 
	
On the right  in the blue histogram we evaluate the probability density modelled by the kernel in Eq.(\ref{omega}) in presence of the same configuration of elements. The same trend is obtained considering the probability density distribution, as visualized in Figure \ref{hist} (right). In order to consider the inhibitory effect we evaluate the kernel with 0 mean. A quantitative analysis of the differences between them have been evaluated considering the mean square error between the two normalized histograms. The error of $8\%$ underlines how this connectivity kernel well represents neural connections.

\section{Emergence of percepts}

In the following experiments some numerical simulations will be performed in order to test the reliability of the method for 
performing grouping and detection of perceptual units in an image. 
The kernel considered here only depends on orientation. Hence it can be applied 
			to detect the saliency of geometrical figures 
			which can be very well described using this feature.\\

The purpose is to  select the perceptual units in these images, using the following algorithm: 
\\

\textit{1. Define the affinity matrix $A_{i,j}$ from the connectivity kernel. \\
	\mbox{\quad} 2. Solve the eigenvalue problem $A_{i,j}$$u_i$ = $\lambda_i$$u_i$, where the order of $i$ is such that $\lambda_i$ is decreasing. \\
	\mbox{\quad} 3. Find and project  on the segments the eigenvector $u_1$ associated to the largest eigenvalue.}\\

The parameters used are: 1000000 random paths with $\sigma = 0.15$ in the system (\ref{system_FP}), $\sigma_1 =  1.2$, $\sigma_3 = 0.11$ in the system (\ref{system_SRL}), $\sigma$, $\rho = 0.15$ in the system (\ref{system_L}). The value of $H$ is defined as follows: $H = \frac{1}{3}d_{max}$, where $d_{max}$ is the maximum distance between the inducers of the stimulus. Similar parameters have been used for all the experiments.

\subsection{The Field, Hayes and Hess experiment}			

\begin{figure}[H]
		\centering 
		\begin{subfigure}[b]{1.2 in} \centering
	\includegraphics[trim = 0.2in 0.6in 0.35in 0.3in, clip, width=1.4in]{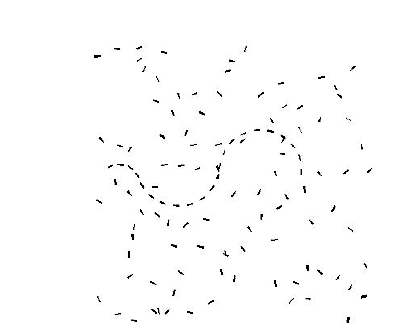}
	 	\caption{}
		\end{subfigure}
		\begin{subfigure}[b]{1.5 in} \centering
		\includegraphics[trim = 0.2in 0.3in 0.2in 0.2in, clip, width=1.4in]{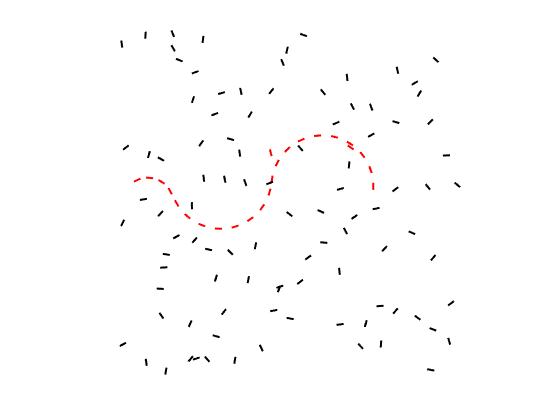}
	 	\caption{}
		\end{subfigure}\\
						 
						 \begin{subfigure}[b]{1.1 in} \centering
						 	\includegraphics[width=1.2\textwidth]{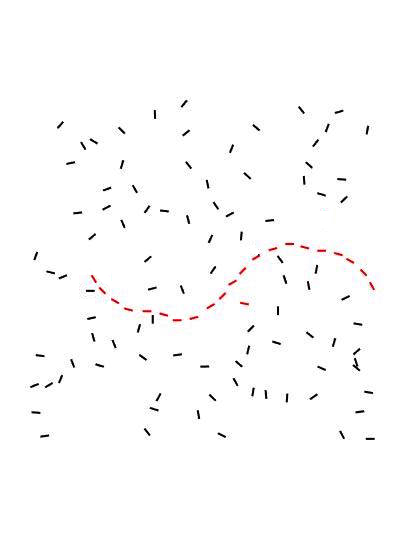}
						 	\caption{}
						 \end{subfigure}
						 \begin{subfigure}[b]{1.1 in} \centering
						 	\includegraphics[width=1.2\textwidth]{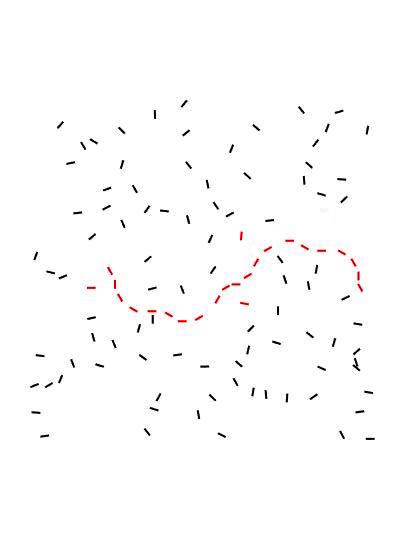}
						 	\caption{} 
						 \end{subfigure}
						 \begin{subfigure}[b]{1.1 in} \centering
						 	\includegraphics[width=1.2\textwidth]{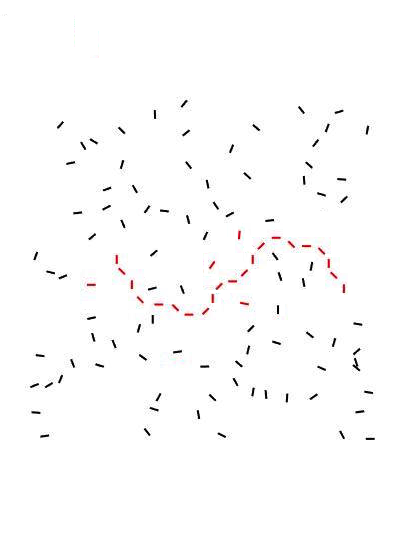}
						 	\caption{}
						 \end{subfigure}
						 \begin{subfigure}[b]{1.1 in} \centering
						 	\includegraphics[width=1.2\textwidth]{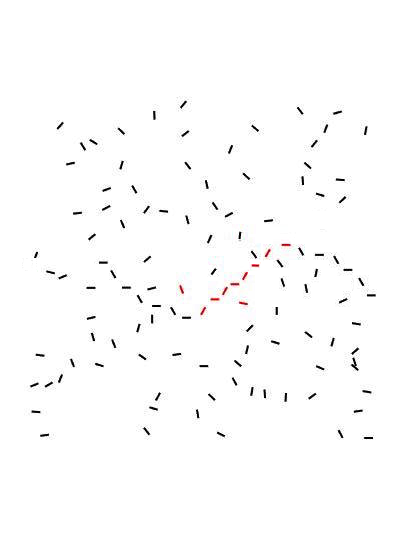}
						 	\caption{}
						 \end{subfigure}
						 \begin{subfigure}[b]{1.1 in} \centering
						 	\includegraphics[width=1.2\textwidth]{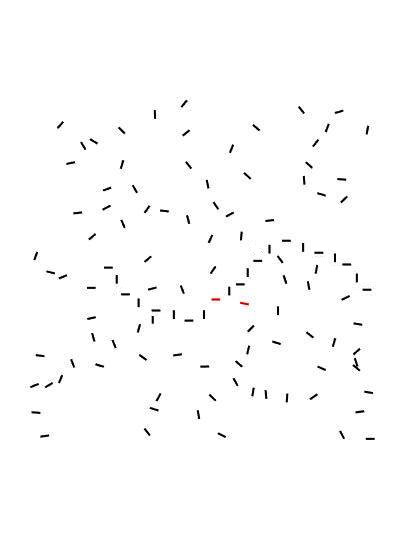}
						 	\caption{}
						 \end{subfigure}
		\begin{subfigure}[b]{1.2 in} \centering
			\includegraphics[trim = 0.4in 0.8in 0.3in 0.2in, clip, width=1.4in]{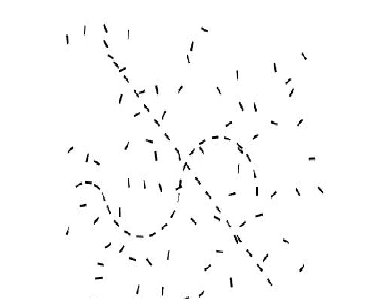}
		\end{subfigure}
		\begin{subfigure}[b]{1.2 in} \centering
			\includegraphics[trim = 0.4in 0.8in 0.3in 0.2in, clip, width=1.4in]{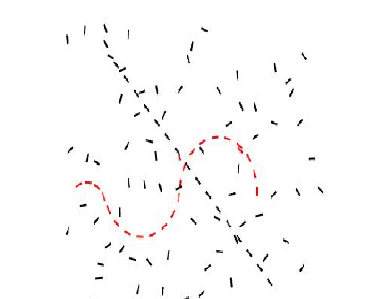}
		\end{subfigure}
		\begin{subfigure}[b]{1.2 in} \centering
			\includegraphics[trim = 0.4in 0.8in 0.3in 0.2in, clip, width=1.4in]{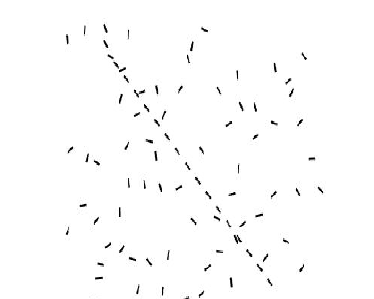}
		\end{subfigure}
		\begin{subfigure}[b]{1.2 in} \centering
			\includegraphics[trim = 0.4in 0.8in 0.3in 0.2in, clip, width=1.4in]{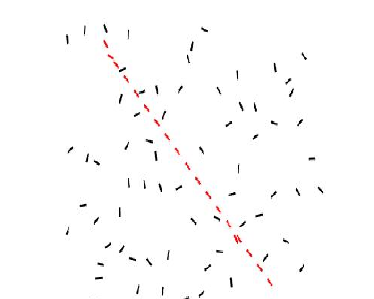}
		\end{subfigure}\\
		\begin{subfigure}[b]{1.2 in} \centering
			\includegraphics[trim = 0.4in 0.8in 0.3in 0.2in, clip, width=1.4in]{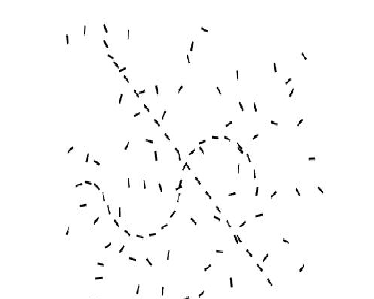}
			\caption{}
			\end{subfigure}
		\begin{subfigure}[b]{1.2 in} \centering
			\includegraphics[trim = 0.4in 0.8in 0.3in 0.2in, clip, width=1.4in]{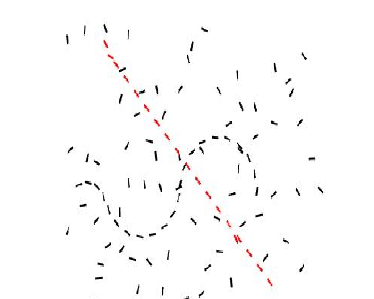}
			\caption{} 
		\end{subfigure}
		\begin{subfigure}[b]{1.2 in} \centering
			\includegraphics[trim = 0.4in 0.8in 0.3in 0.2in, clip, width=1.4in]{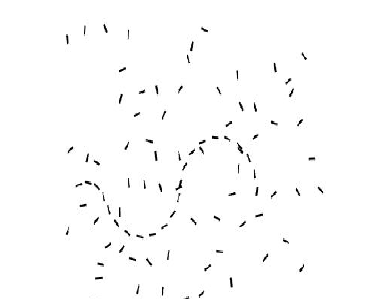}
			\caption{}
			\end{subfigure}
		\begin{subfigure}[b]{1.2 in} \centering
			\includegraphics[trim = 0.4in 0.8in 0.3in 0.2in, clip, width=1.4in]{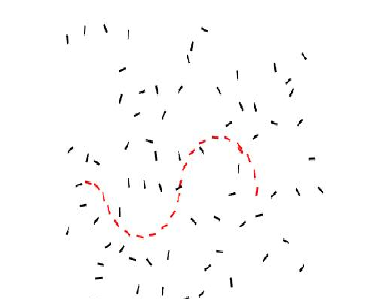}
			\caption{}
		\end{subfigure}

	\caption[Perceptual Unit]{First row. Example of stimulus (a) similar to the experiments of \citep{field1993contour}. The stimulus containing a perceptual unit is segmented with Fokker Planck and Sub-Riemannian Laplacian (b), using the first eigenvector of the affinity matrix. \\ Second row.
		In red the first eigenvector of the affinity matrix considering images containing paths in which the orientation of successive elements differs by 15 (c), 30 (d), 45 (e), 60 (f) and 90 (g) degrees.\\
		Third and fourth rows. Examples (h) with two units in the scene, where a change in the agle leads to a change in the order of the eigenvalues (i),(j),(k). \\ }
	\label{Per}
\end{figure}

In this section we consider some experiments similar to the ones of Field, Hayes and Hess \citep{field1993contour}, where a subset of elements organized in a coherent way is presented out of a ground formed by a random distribution of elements.  
A first stimulus of this type is represented in Figure \ref{Per} (first row). 
The connectivity among these elements is defined as in equations (\ref{GammaFP}) and (\ref{GammaSRL}). 

After the affinity matrix and its eigenvalues, the eigenvector corresponding to the highest eigenvalue is visualized in red.
The results show that the stimulus is well segmented with the fundamental solutions of Fokker Planck and Sub-Riemannian Laplacian equations (Figure \ref{Per} (b)).

 Now we consider a similar experiment proposed in \citep{field1993contour},  where the orientation of successive elements differs by 15, 30, 45, 60 and 90 degrees and the ability of the observer 
to detect the path was measured experimentally. It was proved 
that the path can be identified when the successive elements differ by 60 deg or less. With our method, we obtain similar results: if the angle between successive elements is less than 60 degree (Figure \ref{Per} (c), (d),(e)), 
the identification of the unit is correctly performed. With an angle equal to 60 degree (Figure \ref{Per} (f)) only a part of the curve is correctly detected: this can be interpreted as the increasing observer's difficulty to detect the path. Finally with higher angles (Figure \ref{Per} (g)) the first eigenvector of the affinity matrix corresponds to random inducers, confirming the results.  
\\
Finally we present an example where there are two units in the scene with roughly-equal salience, 
they have roughly-equal eigenvalues. 
		In the third and in the fourth row of Figure \ref{Per} the stimuli are  composed by a curve 
		and a line in a background of random elements. 
		In the stimulus (h) represented in the third row, 
		the elements composing the curve are perfectly aligned and very nearby, so that this 
		has the highest saliency and it represents the eigenvector associated to the first eigenvalue (as shown in red in Figure \ref{Per} (i)). The second eigenvalue in this case is sligtly smaller. 
		After the computation of the first eigenvector, the stimulus is updated  (Figure \ref{Per} (j)), the first eigenvector of the new affinity matrix is computed and it corresponds to the inducers of the line (Figure \ref{Per} (k)). 
		
		In the fourth row we slightly modify the stimuli, in particular the alignement of the element forming the curve (e.g. an angle of pi/18). 
		As a consequence, the line becomes the most salient perceptual unit and the first eigenvector 
		 (Figure \ref{Per} (i), fourth row). The stimulus is updated  (Figure \ref{Per} (j), fourth row) and the first eigenvector of the new affinity matrix corresponds to the inducers of the curve (Figure \ref{Per} (k), fourth row). 
		 It is notable that in this case a small changement of the eigenvalues corresponds to small changement of the eigenvectors, 
		but the first eigenvalue swaps with the second one and consequently we obtain an abrupt change in the perceved object.\\

\subsection{The role of polarity}
	
The term of polarity leads to insert in the model the feature of contrast: contours with the same orientation but opposite contrast are referred to opposite angles.
	For this reason we assume that the orientation $\theta$ takes values in $[0,2\pi $) when we consider the odd filters and in $[0,\pi $) while studying the even ones.
	\begin{figure}[H]
		\centering
		\includegraphics[width=0.18\textwidth]{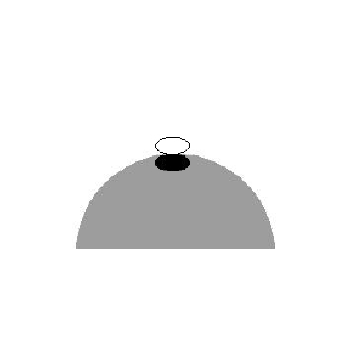}	
		\includegraphics[width=0.18\textwidth]{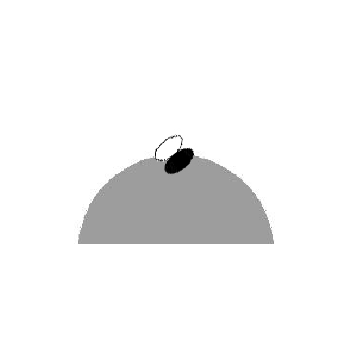}	
		\includegraphics[width=0.18\textwidth]{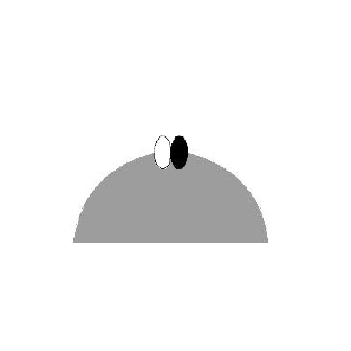}	
		\includegraphics[width=0.18\textwidth]{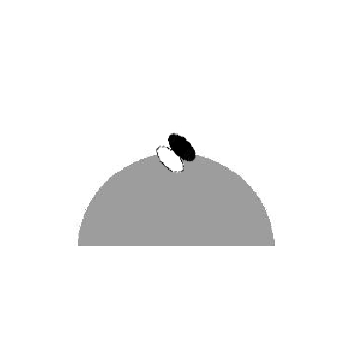}
		\includegraphics[width=0.18\textwidth]{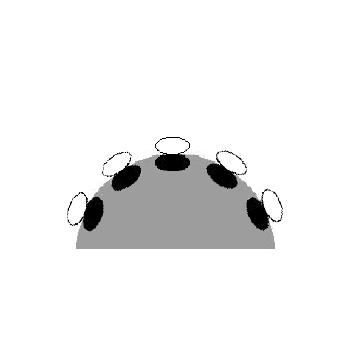}\\
		\begin{subfigure}[b]{1.1 in} \centering
			\includegraphics[width=1.2\textwidth]{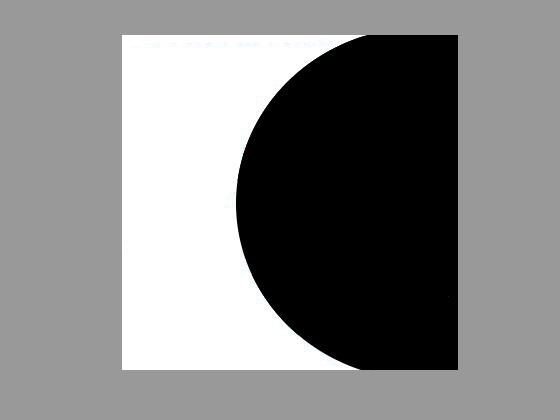}
			\caption{}
			\end{subfigure}
	\begin{subfigure}[b]{1.1 in} \centering
		\includegraphics[width=1.2\textwidth]{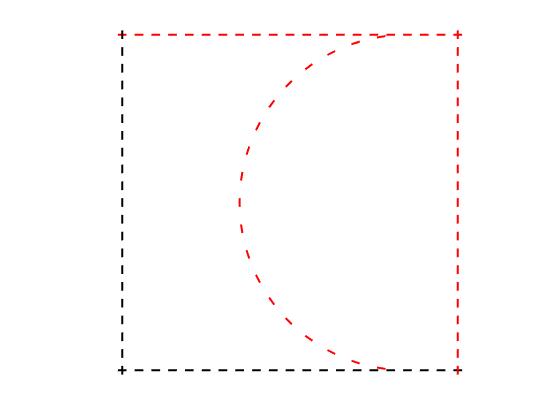}
		\caption{}
		\end{subfigure}
			\begin{subfigure}[b]{1.1 in} \centering
				\includegraphics[width=1.2\textwidth]{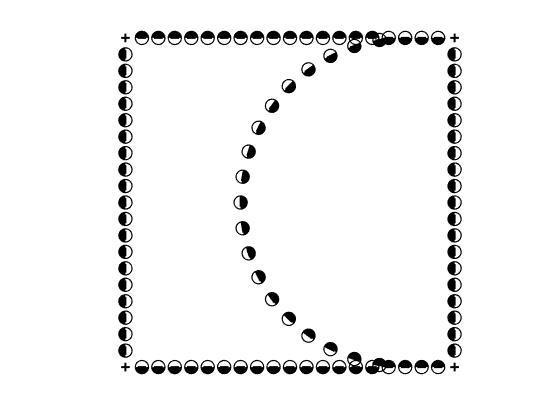}
				\caption{}
	\end{subfigure}
	\begin{subfigure}[b]{1.1 in} \centering
		\includegraphics[width=1.2\textwidth]{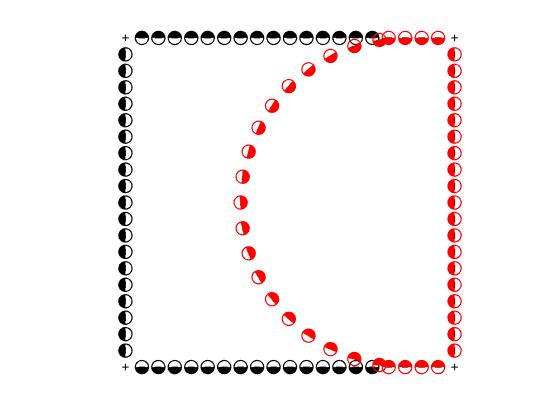}
		\caption{}
	\end{subfigure}		
		\caption[Odd]{In the first row schematic description of the whole hypercolumn of odd simple cells centered in a point $(x,y)$. The maximal activity is observed for the simple cell sensitive to the direction of the boundary of the visual stimulus. The set of maximally firing cells are visualized in the last image. \\ In the second row: 
					a cartoon image (a), the first eigenvector of the affinity matrix without polarity (b), its representation with polarity dependent Gabor patches (b) and the corresponding first eigenvector (d).}  \label{G_odd}
	\end{figure}
	
The response of the odd filters in presence of a cartoon image is schematically represented in Figure \ref{G_odd}. At every boundary point the maximally activated cell is the one with the same direction of the boundary. Then the maximally firing cells are aligned with the boundary (Figure \ref{G_odd}, top right). 

In order to clarify the role of polarity we  consider an image in Figure \ref{G_odd} (a), that has been studied by \citep{kanizsa1980}, in the contest of a study of convexity in perception. 
 In this case, if we consider only orientation of the boundaries 
without polarity, we completely loose any contrast information and 
we obtain the grouping in figure \ref{G_odd} (b). Here  the upper edge of the square 
is grouped as an unique perceptual unit. 
On the other side, while 
inserting polarity, the Gabor patches 
on the upper edge boundary of the black or white region 
have opposite contrast and  detect values of $\theta$ 
which differs of $\pi$ (see Figure \ref{G_odd} (c)). In this way, there is no affinity between these patches, 
and the first eigenvector of the affinity matrix 
represented in red correctly detects the unit present in the image and corresponds to the inducers of the semicircle 
(see Figure \ref{G_odd} (d)).  
 This underlines the important role of polarity in perceptual individuation and segmentation. 
 We also note that the fist perceptual unit detected is the convex one, as 
predicted by the gestalt law (see \citep{kanizsa1980}).

\subsection{The Kanizsa illusory figures}  		
			
We consider here stimuli formed by Kanizsa figures, represented by oriented segments that simulate the output of simple cells. In \citep{lee2001} it is described that completion of Kanizsa figures takes place in V1.

\begin{figure}[H]		\centering 
		\begin{subfigure}[b]{1.1 in} \centering
			\includegraphics[width=1.2\textwidth]{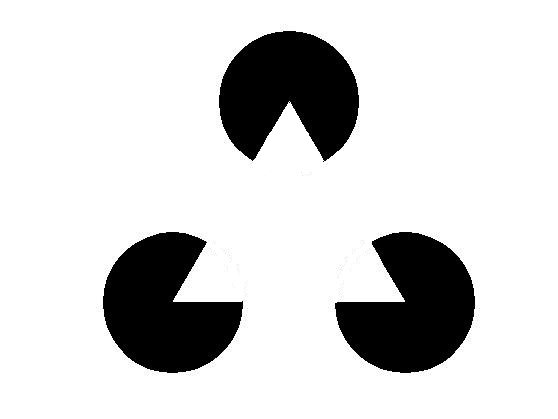}
			\caption{}
		\end{subfigure}
		\begin{subfigure}[b]{1.1 in} \centering
			\includegraphics[width=1.2\textwidth]{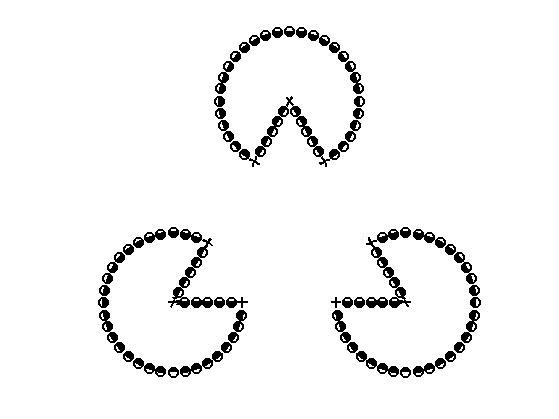}
			\caption{}
		\end{subfigure}
		\begin{subfigure}[b]{1.1 in} \centering
			\includegraphics[width=1.2\textwidth]{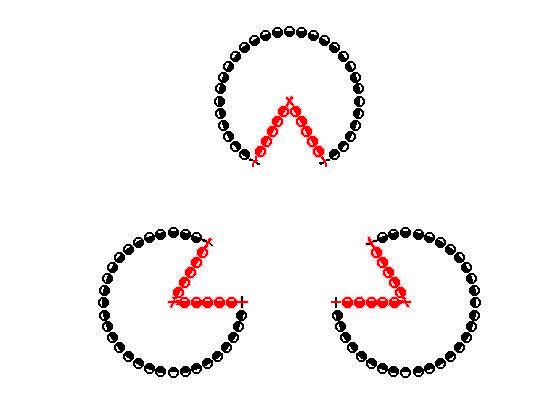}
			\caption{}
		\end{subfigure}
		\begin{subfigure}[b]{1.1 in} \centering
			\includegraphics[width=1.2\textwidth]{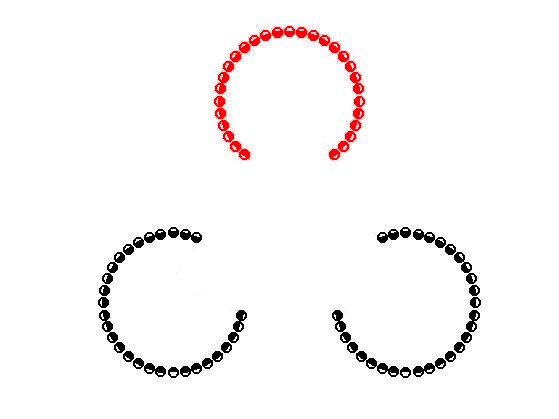}
			\caption{}
		\end{subfigure}
				\caption[Eigenvector]{The Kanizsa triangle (a) and the maximally responding odd filters (b). In (c) it is shown the first eigenvector of the affinity matrix, using the fundamental solutions of Fokker Planck (\ref{GammaFP},\ref{system_FP}) and Sub-Riemannian Laplacian equations (\ref{GammaSRL},\ref{system_SRL}). After this computation, the affinity matrix is updated removing the detected perceptual unit; the first eigenvector of the new matrix is visualized (d). 
					} 
					\label{firstexp}
\end{figure}

We first consider the stimulus of Figure \ref{firstexp} (a). The connectivity among its elements will be analysied with the kernels defined in equations Eq. (\ref{GammaFP}),(\ref{GammaSRL}).

\begin{figure}[H]
			\centering 
				\begin{subfigure}[b]{1.1 in}  
					\includegraphics[width=1.2\textwidth]{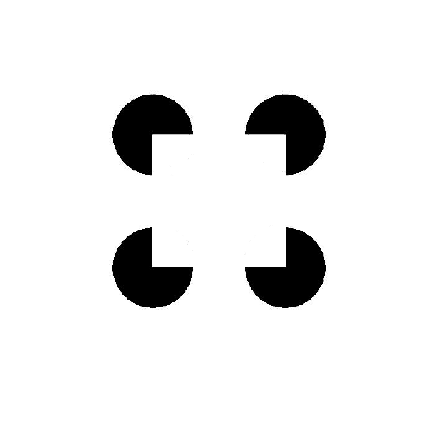}
				\end{subfigure}
				\begin{subfigure}[b]{1.1 in}  
					\includegraphics[width=1.2\textwidth]{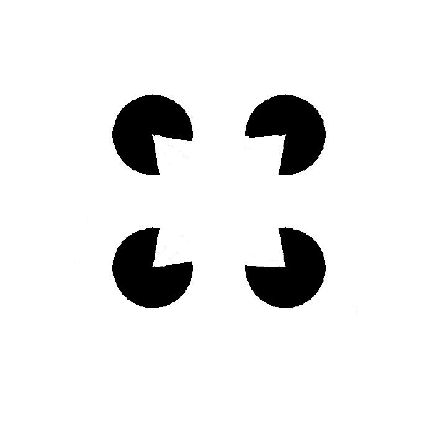}
				\end{subfigure}
				\begin{subfigure}[b]{1.1 in}  
					\includegraphics[width=1.2\textwidth]{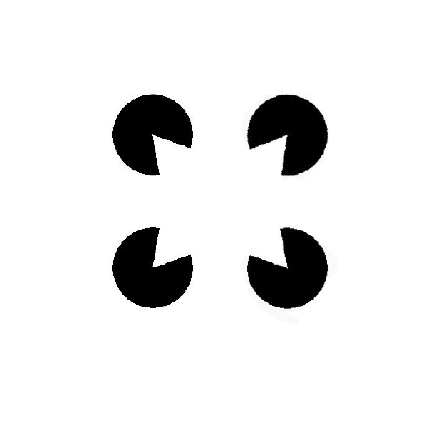}
				\end{subfigure}
				\begin{subfigure}[b]{1.1 in}  
					\includegraphics[width=1.2\textwidth]{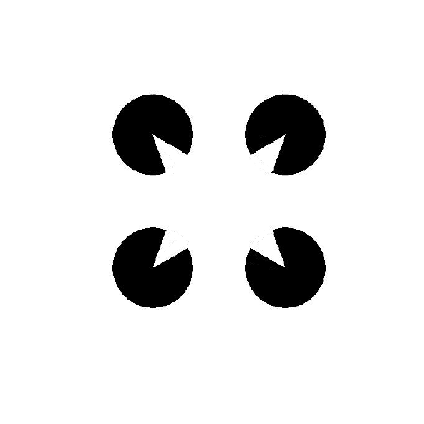}
				\end{subfigure}
				\begin{subfigure}[b]{1.1 in}  
					\includegraphics[width=1.2\textwidth]{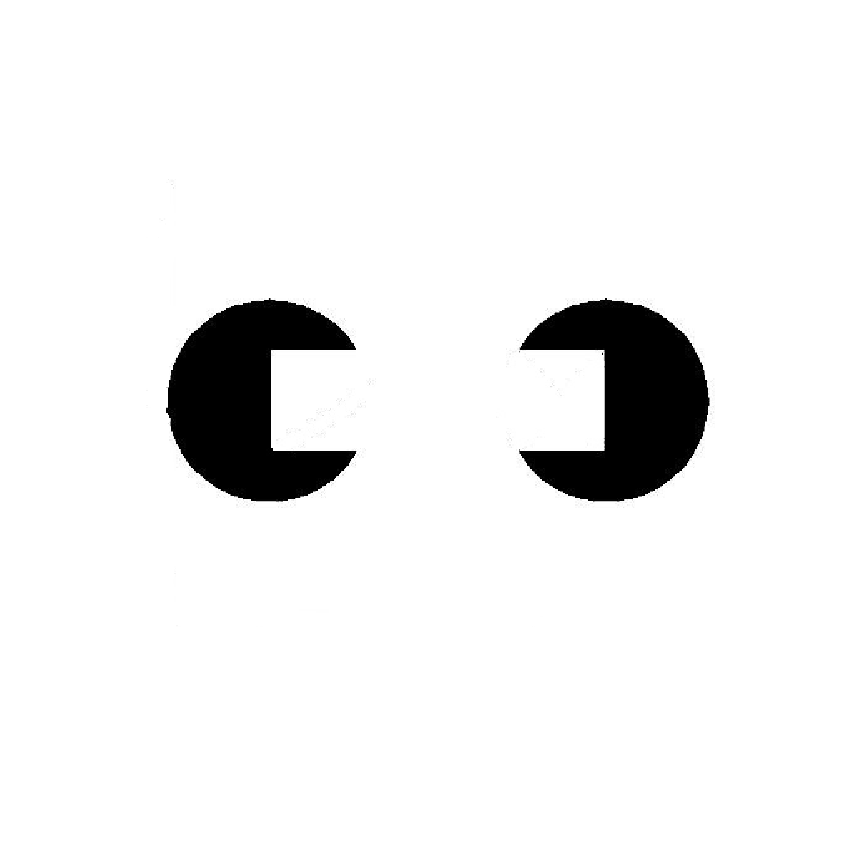}
				\end{subfigure}				
				\begin{subfigure}[b]{1.1 in}  
					\includegraphics[width=1.2\textwidth]{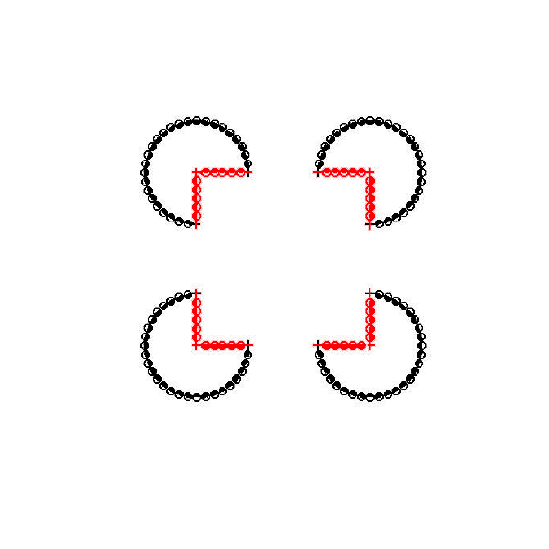}
						\caption{}
				\end{subfigure}
				\begin{subfigure}[b]{1.1 in}  
					\includegraphics[width=1.2\textwidth]{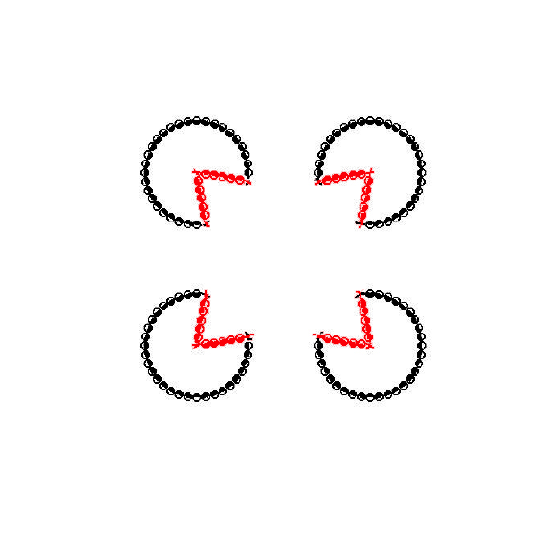}
						\caption{}
				\end{subfigure}
				\begin{subfigure}[b]{1.1 in}  
					\includegraphics[width=1.2\textwidth]{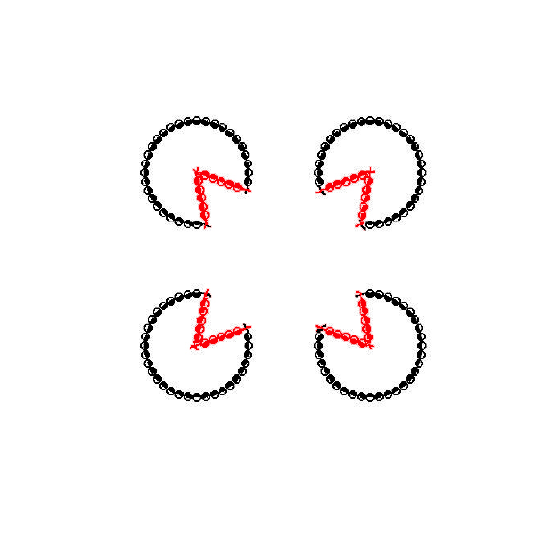}
						\caption{}
				\end{subfigure}
				\begin{subfigure}[b]{1.1 in}  
					\includegraphics[width=1.2\textwidth]{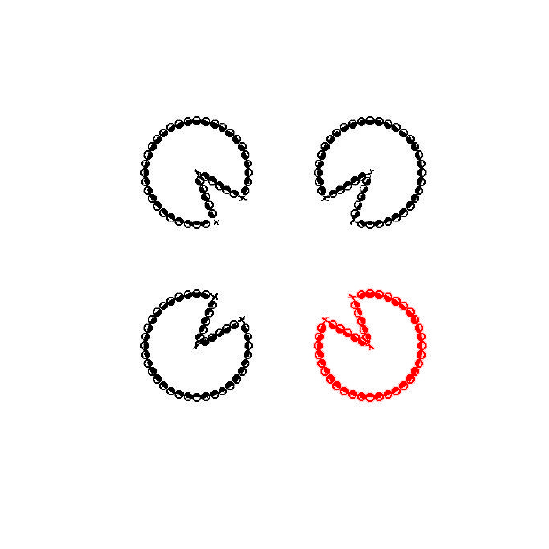}
						\caption{}
				\end{subfigure}
				\begin{subfigure}[b]{1.1 in}  
					\includegraphics[width=1.2\textwidth]{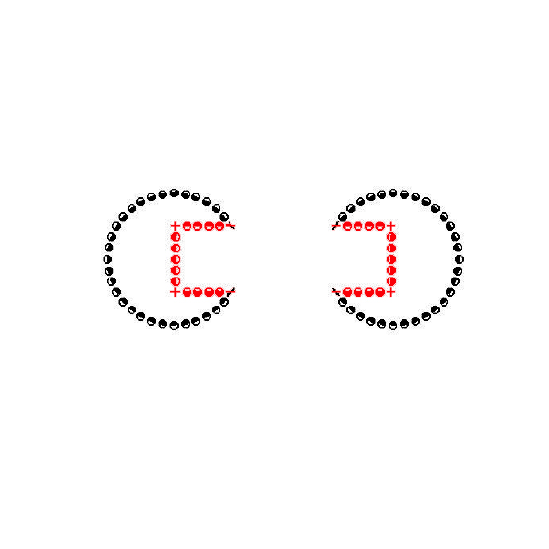}
					\caption{}
				\end{subfigure}
				\begin{subfigure}[b]{1.1 in}  
					\includegraphics[width=1.2\textwidth]{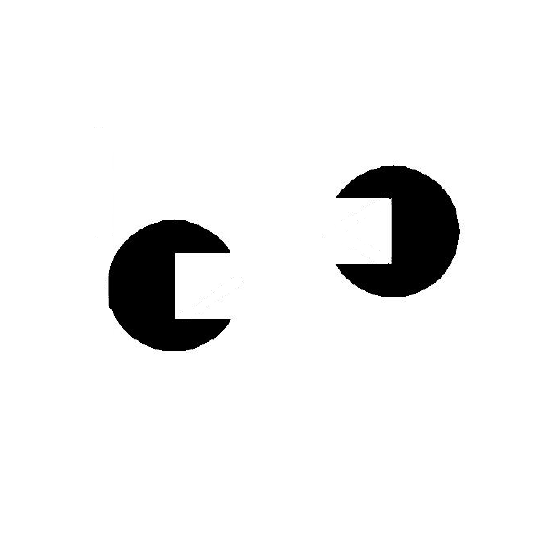}
				\end{subfigure}
				\begin{subfigure}[b]{1.1 in}  
					\includegraphics[width=1.2\textwidth]{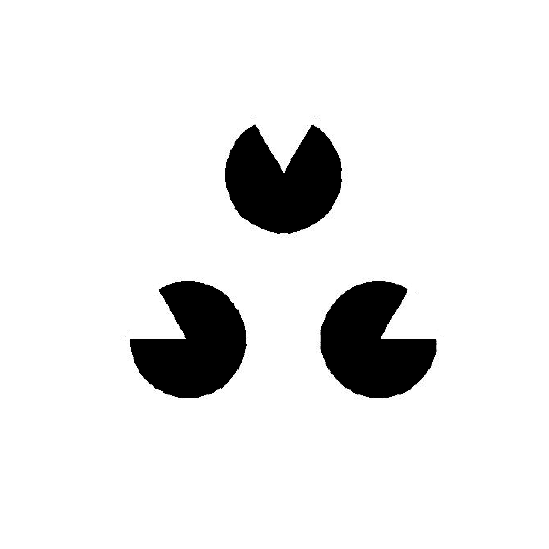}
				\end{subfigure}
				\begin{subfigure}[b]{1.1 in}  
					\includegraphics[width=1.2\textwidth]{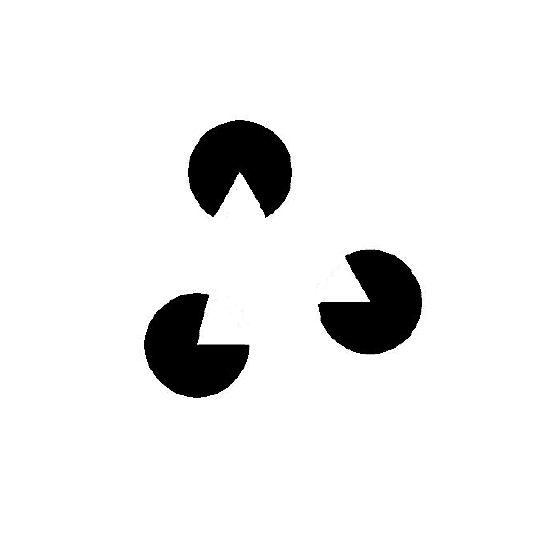}
				\end{subfigure}

				\begin{subfigure}[b]{1.1 in}  
					\includegraphics[width=1.2\textwidth]{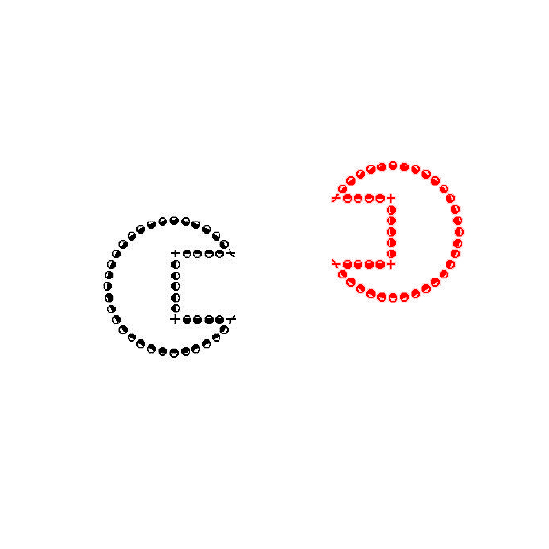}
					\caption{}
				\end{subfigure}
				\begin{subfigure}[b]{1.1 in}  
					\includegraphics[width=1.2\textwidth]{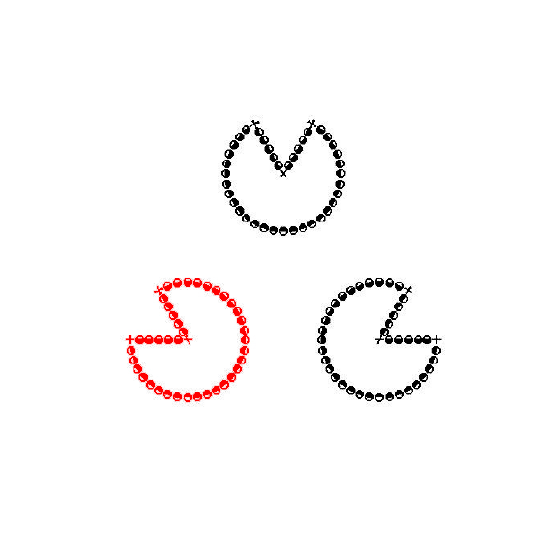}
					\caption{}
				\end{subfigure}
				\begin{subfigure}[b]{1.1 in}  
					\includegraphics[width=1.2\textwidth]{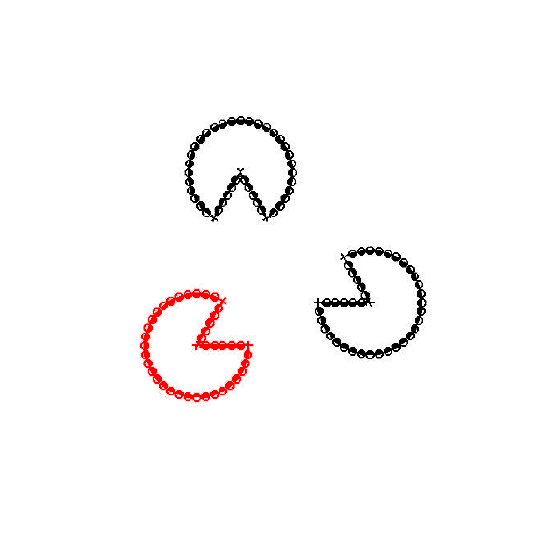}
					\caption{}
				\end{subfigure}
	\caption[Eigenvector]{Examples of stimulus (first row) with aligned and not-aligned inducers.  
		Stimulus with rotated (g) and not-aligned (f),(h) inducers (third row). The first eigenvectors of the affinity matrix using the fundamental solutions of Fokker Planck and Sub-Riemannian Laplacian are visualized in red (second and fourth row).} 
			\label{figure_square}
\end{figure}

The results of simulations with the fundamental solutions of Fokker Planck and Sub-Riemannian Laplacian equations are shown in Figure \ref{firstexp}. The first eigenvector is visualized in red and it 
corresponds to the inducers of the Kanizsa triangle (Figure \ref{firstexp} (c)). In this example, after the computation of the first eigenvector of the affinity matrix, this matrix is updated removing the identified perceptual unit and then the first eigenvector of the new matrix is computed (Figure \ref{firstexp} (d))): these simulations show that circles are 
associated to the less salient eigenvectors. In that way, 
the first eigenvalue can be considered as a quantitative measure of saliency, because it allows to segment the most important object in the scene and the results of simulations confirm the visual grouping. %
  When the affinity matrix is formed by different eigenvectors with almost the same eigenvalues, as in Figure \ref{firstexp} (d), it is not possibile to recognize a most salient object, because they all have the same influence.
 We choose here to show just one inducer in red. The other two have the same eigenvalue. That also happens, for example, when the inducers are not co-circularly aligned or they are rotated.\\
Now we consider as stimulus the Kanizsa square and then we change the angle between the inducers, so that the subjective contours become curved (Figure \ref{figure_square} (a), (b), (c), (d), first row). The fact that illusory figures are perceived depends on a limit curvature. Indeed we perceive a square in the first three cases, but not in the last one.
The results of simulations with the fundamental solutions of Fokker Planck and Sub-Riemannian Laplacian equations confirm the visual grouping (Figure \ref{figure_square} (a), (b), (c), (d), second row): when the angle between the inducers is not too high (cases (a), (b), (c)) the first eigenvector corresponds to the inducers that form the square, otherwise (case (d)) the pacman becomes the most salient objects in the image. In this case, we obtain 4 eigenvectors with almost the same eigenvalue. \\
Now we consider a Kanizsa bar (Figure \ref{figure_square} (e), first row), that is perceived only if the inducers are aligned. Also in that case, the result of simulation confirms the visual perception if we use the fundamental solutions of the Fokker-Planck and the Sub-Riemannian Laplacian equations. When the inducers are not aligned, all the kernels confirm the visual perception, showing two different perceptual units (Figure \ref{figure_square} (f)).

Considering a stimulus composed of rotated or not-aligned inducers, as in Figure \ref{figure_square} (g), (h) it is not possible to perceive it and the results of simulations, using all the connectivity kernels described, confirm the visual grouping. In that case, the affinity matrix is decomposed in 3 eigenvectors with almost the same eigenvalues, which represent the 3 perceptual units in the scene.

	\subsection{Sub-Riemannian Fokker Planck versus Sub-Riemannian Laplacian}
	
\begin{figure}[!h]
\centering
\begin{subfigure}[b]{1.1 in}  
	\includegraphics[width=1.2\textwidth]{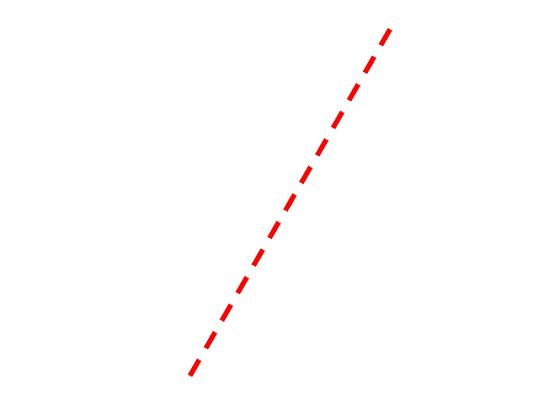}
		\caption{}
\end{subfigure}
\begin{subfigure}[b]{1.1 in}  
	\includegraphics[width=1.2\textwidth]{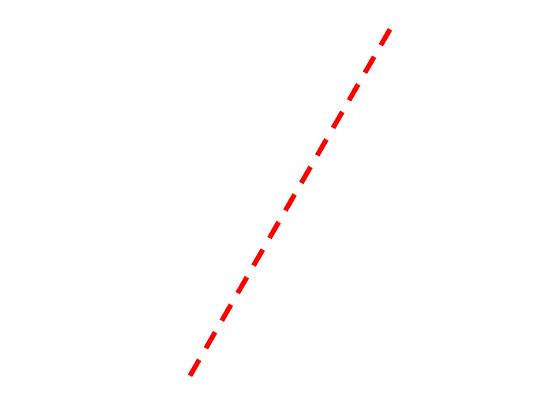}
		\caption{}
\end{subfigure}
\begin{subfigure}[b]{1.1 in}  
	\includegraphics[width=1.2\textwidth]{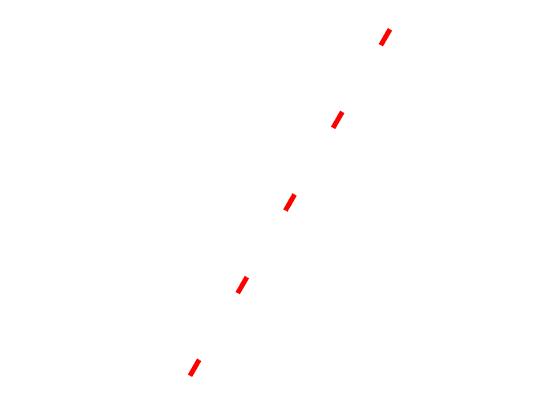}
		\caption{}
\end{subfigure}
\begin{subfigure}[b]{1.1 in}  
	\includegraphics[width=1.2\textwidth]{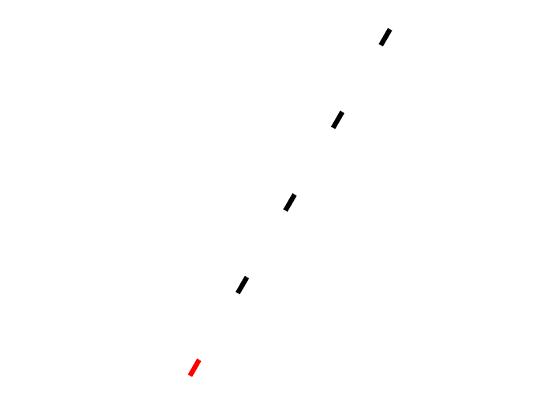}
		\caption{}
	
\end{subfigure}
\\
\begin{subfigure}[b]{1.1 in}  
	\includegraphics[width=1.2\textwidth]{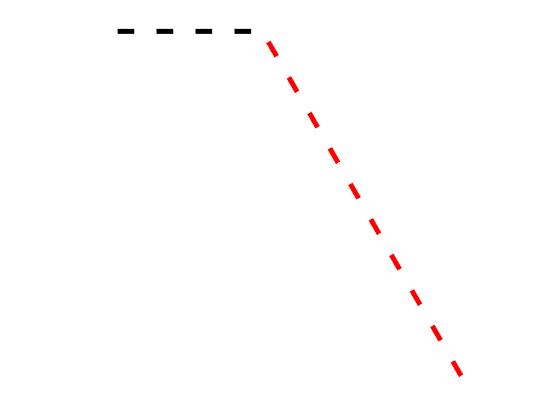}
		\caption{}
	
\end{subfigure}
\begin{subfigure}[b]{1.1 in}  
	\includegraphics[width=1.2\textwidth]{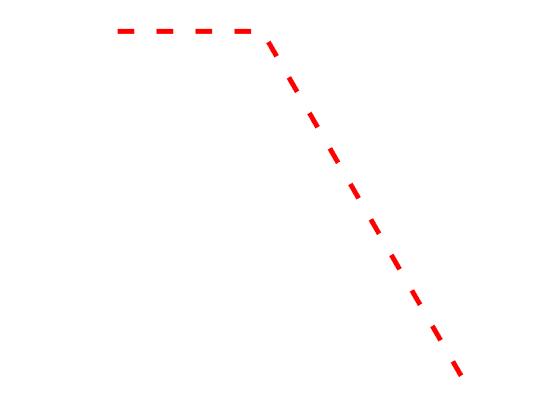}
		\caption{}
\end{subfigure}
\\
\caption{In the first row a few aligned segments, which are 
correctly grouped by the Fokker Planck and the Sub-Riemannian Laplacian (a), (b). 
When we separate the inducers, the perceptual unit is correctly 
detected using the Fokker Planck kernel (c), while the Sub-Riemannian Laplacian is 
not able to perform the grouping (d). In the second row we consider an angle. 
In this case the Fokker Planck is unable to perform the grouping (e), 
while the Sub-Riemannian Laplacian can correctly perform the grouping (f). 
}\label{FPSL}
\end{figure}

We have outlined in Section 2.2 and 3.2 that the Fokker Planck kernel accounts for long range connectivity, while Sub-Riemannian Laplacian for short range. In the previous examples we obtain good results with both kernels, but this difference emerges while we suitable change the parameters. 
In Figure \ref{FPSL} we compare the action of these two kernels.

 In the first row we see some segments, which form an unique perceptual unit. 
 If they are not too far, the grouping is correctly performed both 
 by the Fokker Planck and the Sub-Riemannian Laplacian (Figure \ref{FPSL} (a),(b)). 
 When we separate the inducers, the perceptual unit is correctly 
 detected by the Fokker Planck kernel (Figure \ref{FPSL} (c)), while the Sub-Riemannian Laplacian is 
 not able to perform the grouping  (Figure \ref{FPSL} (d)). This confirms that the Fokker Planck kernel is responsible 
 for long range connectivity. 
 In the second row we consider an angle. When the angle is sufficiently big, 
 the Fokker Planck becomes unable to perform the grouping (Figure \ref{FPSL} (e)), 
 while the Sub-Riemannian Laplacian, correctly performs the grouping 
 of the perceptual unit (Figure \ref{FPSL} (f)). This confirms that the Sub-Riemannian Laplacian 
 can be used as a model of short range connectivity.

	\subsection{Sub-Riemannian versus Riemannian kernels}

\begin{figure}[H]
			\centering 
				\includegraphics[width=.17\textwidth]{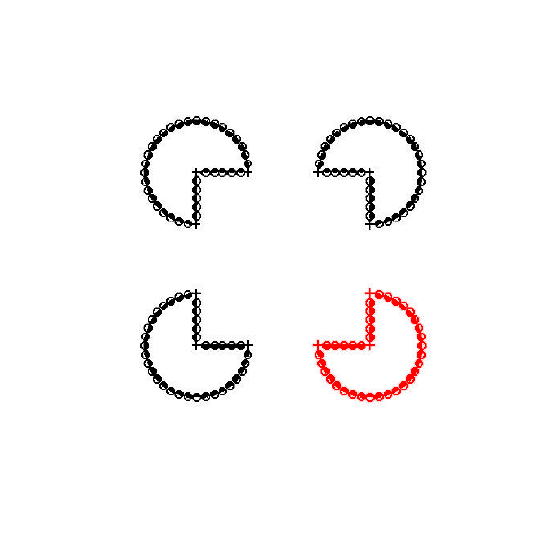}	
				\includegraphics[width=.17\textwidth]{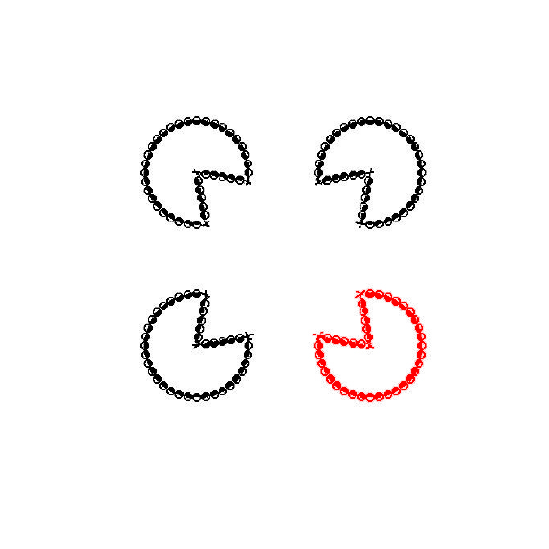}
				\includegraphics[width=.17\textwidth]{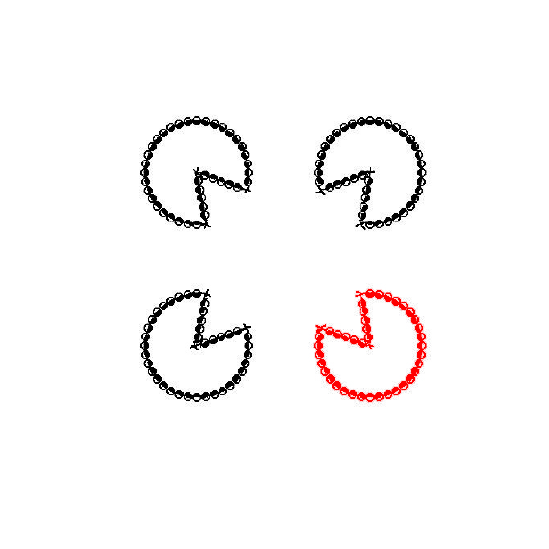}	
				\includegraphics[width=.17\textwidth]{quad_u4.pdf} \\
				\includegraphics[width=0.17\textwidth]{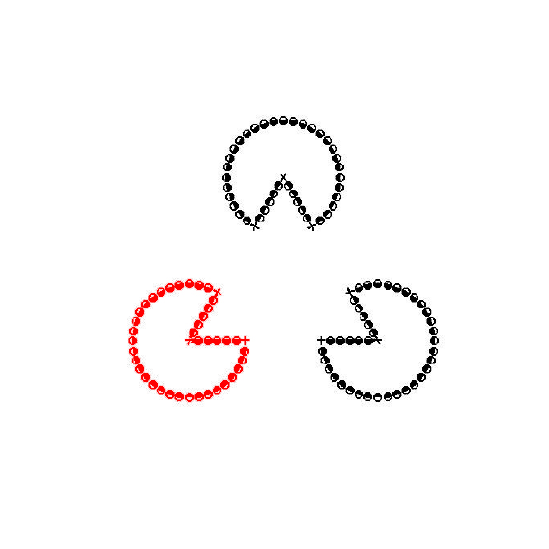}
				\includegraphics[width=0.17\textwidth]{Kround_R.pdf}
				\includegraphics[width=0.17\textwidth]{Ksfas_R.pdf}
				\includegraphics[width=.17\textwidth]{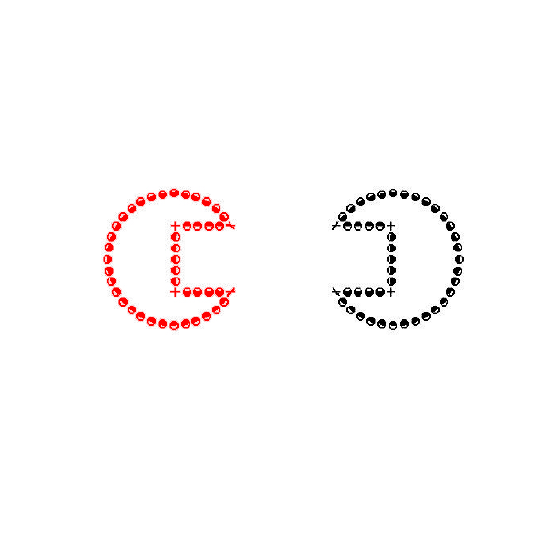}
				\includegraphics[width=.17\textwidth]{asta_sf_R.pdf}			 
				\caption{ Stimulus of Figure \ref{firstexp} and \ref{figure_square}. The results do not fit the visual perception if we use the isotropic Laplacian equation (\ref{GammaISO} , \ref{system_L}) and confirm the necessity to use a Sub-Riemannian kernel to model the 
				cortical connectivity.} \label{lap}
\end{figure}

In order to further validate the Sub-Riemannian model we show that the 
model applied with the isotropic Laplacian kernel 
does not perform correctly. As shown in Figure \ref{lap} (first row)
the visual perception is not correctly modeled: 
the first eigenvectors coincide with one of the inducers 
and the squares are not recognized.
 That also happens for the stimulus of Figure 9 (a) and when the inducers are not co-circularly aligned or they are rotated.

\section*{Conclusions}

In this work we have presented a neurally based model for figure-ground segmentation using spectral methods, where segmentation has been performed by computing eigenvectors of affinity matrices.\\
Different connectivity kernels that are compatible with the functional architecture of the primary visual cortex have been used.
We have modelled them as fundamental solution of Fokker-Planck, Sub-Riemannian Laplacian and isotropic Laplacian equations and compared their properties.\\
With this model we have identified perceptual units of different Kanizsa figures, showing that this can be considered a good quantitative model for the constitution of perceptual units equipped by their saliency. We have also shown that the fundamental solutions of Fokker-Planck and Sub-Riemannian Laplacian equations are good models for the good continuation law, while the isotropic Laplacian equation is less representative for this gestalt law. However it retrieves information about ladder parallelism, a feature that can be analysed in the future. All the three kernels are able to accomplish boundary completion with a preference for the operators Fokker Planck and the Sub-Riemannian Laplacian.\\
The proposed mathematical model is then able to integrate local and global gestalt laws as a process implemented in the functional architecture of the visual cortex.
The kernel considered here only depends on orientation. Hence it can be applied 
			to detect the saliency of geometrical figures 
			which can be very well described using this feature. 
			The same method can be applied to natural images if their main features are related to 
			orientations, as in retinal images (see \citep{favali2015}). 
			The ideas presented here could be extended to   more general kernels 
			able to detect geometrical features different from orientation and we are confident that there is a relation between the highest 
			eigenvector and the salient object. 
			However for general images we can not rely on this simple geometric method, 
			since different cortical areas can be involved in the definition of the saliency, 
			with a modulatory effect on the connectivity of V1.

\renewcommand{\abstractname}{Acknowledgements}
\begin{abstract}
The research leading to these results has received funding from the People Programme (Marie Curie Actions) of the European Union's Seventh Framework Programme FP7/2007-2013/ under REA grant agreement n°607643
\end{abstract}


\begin{thebibliography}{100}
	\providecommand{\natexlab}[1]{#1}
	\expandafter\ifx\csname urlstyle\endcsname\relax
	\providecommand{\doi}[1]{doi:\discretionary{}{}{}#1}\else
	\providecommand{\doi}{doi:\discretionary{}{}{}\begingroup
		\urlstyle{rm}\Url}\fi
	
	\bibitem[{Angelucci et~al.(2002) Angelucci, Levitt, \& Walton}]{angelucci2002circuits}
	Angelucci, A., Levitt, J. B., Walton, E. J. S., Hupe, J. M., Bullier, J., \& Lund, J.S.(2002).
	\newblock Circuits for Local and Global Signal Integration in Primary Visual Cortex.
	\newblock \emph{The Journal of Neuroscience}, 22(19):86338646.	

	
	\bibitem[{August \& Zucker(2000) August, J. \& Zucker, S.W.}]{august2000curve}
	August, J., \& Zucker S. W. (2000).
	\newblock The curve indicator random field: Curve organization via edge correlation. In
	\newblock \emph{Perceptual organization for artificial vision systems}, 265--288. Springer.
	

	
	\bibitem[{August \& Zucker(2003)August, J. \& Zucker, S.W.}]{august2003sketches}
	August, J. \& Zucker, S. W. (2003).
	\newblock Sketches with curvature: the curve indicator random field and Markov processes.
	\newblock \emph{Pattern Analysis and Machine Intelligence, IEEE Transactions on}, 25(4): 387--400.	

   \bibitem[{Barbieri(2012)}]{barbieri2012}
	 Barbieri, D., Citti, G., Sanguinetti, G., \& Sarti, A. (2012). 
	\newblock An uncertainty principle underlying the functional architecture of V1. 
	\newblock \emph{Journal of Physiology} 106(5-6):183-193. 

	\bibitem[{Belkin \& Niyogi(2003)Belkin, M., \& Niyogi, P.}]{belkin2003laplacian}
	 Belkin, M., \& Niyogi, P. (2003).
	\newblock Laplacian eigenmaps for dimensionality reduction and data representation.
	\newblock \emph{Neural Computation}, 15(6):1373--1396.

	\bibitem[{Boscain et~al.(2012)Boscain, U., Duplaix, J., Gauthier, J.P., \& Rossi, F.}]{boscain2012anthropomorphic}
	Boscain, U., Duplaix, J., Gauthier, J. P., \& Rossi, F. (2012).
	\newblock Anthropomorphic image reconstruction via hypoelliptic diffusion.
	\newblock \emph{SIAM Journal on Control and Optimization}, 50(3):1309--1336.	


	\bibitem[{Bosking et~al.(1997)Bosking, W., Zhang, Y., Schofield,  B.  \& Fitzpatrick D.}]{bosking1997orientation}
	 Bosking, W., Zhang, Y., Schofield,  B.  \& Fitzpatrick D. (1997).
	\newblock Orientation selectivity and the arrangement of horizontal connections in tree shrew striate cortex.
	\newblock \emph{The Journal of neuroscience}, 17(6):2112--2127.	


	\bibitem[{Bressloff et~al.(2002)Bressloff, P.C., Cowan, J.D., Golubitsky, M., Thomas,  P.J., \&  Wiener M.C.}]{bressloff2002geometric}
	 Bressloff, P. C., Cowan, J. D., Golubitsky, M., Thomas,  P. J., \&  Wiener M. C. (2002).
	\newblock What Geometric Visual Hallucinations Tell Us about the Visual Cortex.
	\newblock \emph{Neural Computation}, 14(3): 473--491.

	
	\bibitem[{Bressloff \& Cowan(2003), Bressloff, P.C., \& Cowan, J. D }]{bressloff2003functional}
	Bressloff, P.C., \& Cowan, J. D. (2003).
		\newblock The functional geometry of local and horizontal connections in a model of V1.
		\newblock \emph{Journal of Physiology-Paris}, 221--236.
		
		
	\bibitem[{Citti \& Sarti(2006)Citti, G., \& Sarti, A.}]{citti2006cortical}
    Citti, G., \& Sarti, A. (2006).
	\newblock A cortical based model of perceptual completion in the roto-translation space.
	\newblock \emph{Journal of Mathematical Imaging and Vision},  24(3):307--326.

	
	\bibitem[{Cocci et~al.(2015)Cocci, G., Barbieri, D., Citti, G., \& Sarti, A.}]{cocci2015cortical}
	Cocci, G., Barbieri, D., Citti, G., \& Sarti, A. (2015).
	\newblock Cortical spatio-temporal dimensionality reduction dor visual grouping.
	\newblock \emph{Neural computation}.


	\bibitem[{Coifman \& Lafon(2006)Coifman, R.R., \& Lafon, S.}]{coifman2006diffusion}
	 Coifman, R. R., \& Lafon, S. (2006).
	\newblock Diffusion maps.
	\newblock \emph{Applied and computational harmonic analysis}, 21(1):5--30.	

	\bibitem[{Daugman(1985)}]{daugman1985uncertainty}
	Daugman, J. (1985).
	\newblock Uncertainty relation for resolution in space, spatial frequency, and orientation optimized by two-dimensional visual cortical filters.
	\newblock \emph{JOSA A}, 2(7): 1160--1169.


	\bibitem[{Duits \& Franken(2009)Duits, R.,\& Franken E.M.}]{duits2009line}
	Duits, R.,\& Franken E. M. (2009).
	\newblock Line Enhancement and Completion via Linear Left Invariant Scale Spaces on SE(2). In
	\newblock \emph{Scale Space and Variational Methods in Computer Vision}, 795 -- 807, Springer.

	\bibitem[{Duits \& van Almsick(2008)Duits, R.,\&  van Almsick, M.}]{duits2008explicit}
	 Duits, R.,\&  van Almsick, M. (2008).
	\newblock The explicit solutions of linear left-invariant second order stochastic evolution equations on the 2D-Euclidean motion group.
	\newblock \emph{Quarterly of Applied Mathematics}, 66(1): 27 -- 67.
	
	\bibitem[{Ermentrout \& Cowan(1980)Ermentrout,G.B.,\& Cowan J.D.}]{ErmentroutCowan}
	Ermentrout,G.B.,\& Cowan J.D.(1980).
\newblock Large scale spatially organized activity in neural nets. 
\newblock \emph {SIAM: SIAM Journal on Applied Mathematics}, 38(1):1--21.
 
 \bibitem[{Favali(2015)}]{favali2015}
 Favali, M., Abbasi-Sureshjani, S., ter Haar Romeny, B.: \& Sarti, A. (2015). 
 \newblock Analysis of Vessel Connectivities in Retinal Images by Cortically Inspired Spectral Clustering. 
 \newblock \emph{in press, Journal of Mathematical Imaging and Vision}.  

	\bibitem[{Field et~al.(1993)Field, D., Hayes, A., \& Hess, R.F.}]{field1993contour}
	Field, D., Hayes, A., \& Hess, R. F. (1993).
	\newblock Contour integration by the human visual system: evidence for a local association field.
	\newblock \emph{Vision Research}, 33(2):173 -- 193.

	\bibitem[{Fregnac et~al.(2010)Fregnac, Y., Carelli, P., Pananceau, M.,  \& Monier, C.}]{fregnac2010stimulus}
	 Fregnac, Y., Carelli, P., Pananceau, M.,  \& Monier, C. (2010).
	\newblock Stimulus-driven Coordination of Cortical Cell Assemblies and Propagation of Gestalt Belief in V1.
	\newblock \emph{Dynamic Coordination in the Brain: From Neurons to Mind}, 169--192.	

	
	\bibitem[{Fregnac \& Shulz(1999)Fregnac, Y., \&  Shulz, D. E.}]{fregnac1999activity}
	Fregnac, Y., \&  Shulz, D. E. (1999).
	\newblock Activity-dependent regulation of receptive field properties of cat area 17 by supervised hebbian learning.
	\newblock \emph{Journal of neurobiology},41(1): 69--82.
	
	\bibitem[{Gilbert et~al.(1996)Gilbert, C. D., Das, A., Ito, M., Kapadia, M., \& Westheimer, G.}]{gilbert}
	Gilbert, C. D., Das, A., Ito, M., Kapadia, M., \& Westheimer, G. (1996).
	 \newblock Spatial integration and cortical dynamics. \newblock \emph{Proceedings of the National Academy of Sciences}, 93(2): 615--622.	
	
	\bibitem[{Grossberg \& Mingolla(1985)Grossberg, S. \& Mingolla, E.}]{grossberg1985neural}
	Grossberg, S. \& Mingolla, E. (1985).
	\newblock Neural dynamics of form perception: boundary completion, illusory figures, and neon color spreading.
	\newblock \emph{Psychological review}, 92(2):173.
	
	\bibitem[{Higham(2001)}]{higham2001algorithmic}
	Higham, D. J. (2001).
	\newblock An algorithmic introduction to numerical simulation of stochastic differential equations.
	\newblock \emph{SIAM review}, 43(3):525--546.	

	
	\bibitem[{Hoffman(1989)}]{hoffman1989visual}
	Hoffman, W. C. (1989).
	\newblock The visual cortex is a contact bundle.
	\newblock \emph{Applied Mathematics and Computation}, 32(2):137--167.


	\bibitem[{Hubel \& Wiesel(1962)Hubel, D.H., \& Wiesel, T.N. }]{hubel1962receptive}
	Hubel, D. H.,  \& Wiesel, T. N. (1962).
	\newblock Receptive fiedls, binocular interaction and functional architecture in the cat's visual cortex.
	\newblock \emph{The journal of physiology}, 160(1):106.


	\bibitem[{Hubel \& Wiesel(1977)Hubel, D.H., \& Wiesel, T.N.}]{hubel1977ferrier}
	Hubel, D. H.,  \& Wiesel, T. N. (1977).
	\newblock Ferrier lecture: Functional architecture of macaque monkey visual cortex.
	\newblock \emph{Proceedings of the Royal Society of London B: Biological Sciences,}, 198(1130):1--59.	

	 \bibitem[{Jerison \& Sanchez-Calle(1986)Jerison, D. S., \& Sanchez-Calle, A.}]{Jerison1987}
    Jerison, D. S., \& Sanchez-Calle, A. (1986).
 	\newblock "Estimates for the heat kernel for a sum of squares of vector fields." 
 	\newblock \emph{Indiana University mathematics journal,} 35.4:835--854.

	 
	 
	\bibitem[{Jones \& Palmer (1987)Jones, J.P.,\& Palmer, L.A.}]{jones1987evaluation}
	Jones, J. P., \& Palmer, L. A. (1987).
	\newblock An evaluation of the two-dimensional Gabor filter model of simple receptive fields in cat striate cortex.
	\newblock \emph{Journal of neurophysiology}, 58(6):1233--1258.

	\bibitem[{Kanizsa(1980)}]{kanizsa1980}	
	Kanizsa G. (1980).	
	\newblock Grammatica del vedere
	\newblock \emph{Il Mulino, Bologna}.

	\bibitem[{Kellman \& Shipley(1991)Kellman, P.J., \& Shipley, T.F.}]{kellman1991theory}
	Kellman, P. J., \& Shipley, T. F. (1991).
	\newblock A theory of visual interpolation in object perception.
	\newblock \emph{Cognitive psychology}, 23(2):141--221.


	\bibitem[{Koch \& Ullman(1985)Koch, C., \& Ullman, S.}]{koch1985}
	Koch, C.,\& Ullman,  S. (1985). 
	\newblock Shifts in selective visual attention: towards the underlying neural circuitry.
	\newblock \emph{Human Neurobiology}, 4:219–-227.

	\bibitem[{Koenderink \& van Doorn(1987)Koenderink, J. J., \& van Doorn, A. J.}]{koenderink1987representation}
	Koenderink, J. J., \& van Doorn, A. J. (1987).
	\newblock Representation of local geometry in the visual system.
	\newblock \emph{Biological cybernetics}, 55(6):367--375.


	\bibitem[{Koflka(1935)}]{koflka1935principles}
	Koflka, K. (1935).
	\newblock Principles of Gestalt Psychology.
	\newblock \emph{New York: Har}.	


	\bibitem[{Kohler(1929)}]{kohler1929gestalt}
	Kohler, W. (1929).
	\newblock Gestalt Psychology.
	\newblock \emph{New York: Liveright}.	

	\bibitem[{Lee \& Nguyen(2001) Lee, T. S., \& Nguyen M.}]{lee2001}
	Lee, T. S., \& Nguyen M.	
	\newblock Dynamics of subjective contour formation in the early visual cortex. 
	\newblock \emph{Proceedings of the National Academy of Sciences} (2001).
	 
	 
	\bibitem[{Lorenceau \& Alais(2001)Lorenceau, J., \& Alais, D.}]{lorenceau2001form}
	Lorenceau, J., \& Alais, D. (2001).  
	\newblock Form constraints in motion binding.
	\newblock \emph{Nature neuroscience}, 4(7):745--751.


	\bibitem[{Meila \& Shi(2001)Meila, M., \& Shi, J.}]{meila2001random}
	Meila, M., \& Shi, J. (2001). 
	\newblock A random walks view of spectral segmentation..
	\newblock \emph{8th International Workshop on Articial Intelligence and Statistics}.

		
	\bibitem[{Merleau-Ponty(1945)Merleau-Ponty, M.}]{merleau1996phenomenology}
	Merleau-Ponty, M.  (1945). Translated in 
	\newblock Phenomenology of perception (1996).
	\newblock \emph{Motilal Banarsidass Publishe}.	


	\bibitem[{Mumford(1994)}]{mumford1994elastica}
	Mumford, D. (1994).
	\newblock Elastica and computer vision.
	\newblock \emph{Algebraic geometry and its applications}, 491--506, Springer.	

	
	\bibitem[{Parent \& Zucker(1989)Parent, P., \& Zucker, S. W.}]{parent1989trace}
	Parent, P., \& Zucker, S. W. (1989).
	\newblock Trace inference, curvature consistency, and curve detection.
	\newblock \emph{Pattern Analysis and Machine Intelligence, IEEE Transactions on}, 11(8):823--839.

	\bibitem[{Perona \& Freeman(1998)Perona, P., \& Freeman, W. }]{perona1998factorization}
	Perona, P., \& Freeman, W. (1998).
	\newblock A factorization approach to grouping.
	\newblock \emph{Computer Vision|ECCV'98}, 655--670, Springer.

	\bibitem[{Petitot(2003)}]{petitot2003neurogeometry}
	Petitot, J. (2003).
	\newblock The neurogeometry of pinwheels as a sub-Riemannian contact structure.
	\newblock \emph{Journal of Physiology-Paris}, 97(2):265--309.

	\bibitem[{Petitot(2008)}]{petitot2008}
	Petitot, J. (2008).
	\newblock Neurogéométrie de la vision. Modeles mathématiques et physiques des architectures fonctionelles. 
	\newblock \emph{Paris: Éd. École Polytech} (2008).

	\bibitem[{Petitot \& Tondut(1999)Petitot, J., \& Tondut, Y. }]{petitot1999vers}
	Petitot, J., \& Tondut, Y. (1999).
	\newblock Vers une neurogeometrie. Fibrations corticales, structures de contact et contours subjectifs modaux.
	\newblock \emph{Mathematiques informatique et sciences humaines}, (145):5--102.
		
	\bibitem[{Pillow \& Nava(2002).}]{pillow2002} 
	Pillow, J., \& Nava, R. (2002). 
	\newblock Perceptual completion across the vertical meridian and the role of early visual cortex.
	\newblock \emph{Neuron}, 33.5: 805-813.


	\bibitem[{Ringach(2002)}]{ringach2002spatial}
	Ringach, D. (2002).
	\newblock Spatial structure and symmetry of simple-cell receptive fields in macaque primary visual cortex.
	\newblock \emph{Journal of neurophysiology}, 88(1):455--463.		


	\bibitem[{Robert \& Casella(2013)Robert, C., \& Casella, G.}]{robert2013monte}
	Robert, C., \& Casella, G. (2013).
	\newblock Monte Carlo statistical methods.
	\newblock \emph{Springer Science \& Business Media.}.	


	\bibitem[{Roweis \& Saul(2000)Roweis, S. T., \& Saul, L. K.}]{roweis2000nonlinear}
	Roweis, S. T., \& Saul, L. K. (2000).
	\newblock Nonlinear dimensionality reduction by locally linear embedding.
	\newblock \emph{Science}, 290(5500):2323--2326.


	\bibitem[{Sanguinetti et~al.(2008)Sanguinetti, G., Citti, G., \& Sarti, A.}]{sanguinetti2008image}
	Sanguinetti, G., Citti, G., \& Sarti, A. (2008).
	\newblock Image completion using a diffusion driven mean curvature flowin a Sub-Riemannian Space.
	\newblock \emph{Int. Conf. on Computer Vision Theory and Applications (VISAPP 2008)},46--53.		



	\bibitem[{Sarti \& Citti(2011)Sarti, A., \& Citti, G.}]{sarti2011origin}
	Sarti, A., \& Citti, G. (2011).
	\newblock On the origin and nature of neurogeometry.
	\newblock \emph{La Nuova Critica}.	


	\bibitem[{Sarti \& Citti(2015)Sarti, A., \& Citti, G.}]{sarti2015constitution}
	Sarti, A., \& Citti, G. (2015).
	\newblock The constitution of visual perceptual units in the functional architecture of V1.
	\newblock \emph{Journal of computational neuroscience}, 38(2):285--300.	


	\bibitem[{Sarti et~al.(2008)Sarti, A., Citti, G., \& Petitot, J.}]{sarti2008symplectic}
	Sarti, A., Citti, G., \& Petitot, J. (2008).
	\newblock The symplectic structure of the primary visual cortex.
	\newblock \emph{Biological Cybernetics}, 98(1):33--48.

	\bibitem[{Sarti \& Piotrowski(2015)Sarti, A., \& Piotrowski, D.}]{sarti2015individuation}
	Sarti, A., \& Piotrowski, D. (2015).
	\newblock Individuation and semiogenesis: An interplay between geometric harmonics and structural morphodynamics.
	\newblock \emph{Morphogenesis and Individuation}, 49--73, Springer.


	\bibitem[{Shi \& Malik(2000)Shi, J., \& Malik, J.}]{shi2000normalized}
	Shi, J., \& Malik, J. (2000).
	\newblock Normalized cuts and image segmentation.
	\newblock \emph{Pattern Analysis and Machine Intelligence, IEEE Transactions on}, 22(8):888--905.	


	\bibitem[{Shipley \& Kellman(1992)Shipley, T. F., \& Kellman, P. J.}]{shipley1992perception}
	Shipley, T. F., \& Kellman, P. J. (1992).
	\newblock Perception of partly occluded objects and illusory figures: Evidence for an identity hypothesis.
	\newblock \emph{Journal of Experimental Psychology: Human Perception and Performance}, 18(1):106.
	
	\bibitem[{Shipley \& Kellman(1994)Shipley, T. F., \& Kellman, P. J.}]{shipley1994spatiotemporal}
	Shipley, T. F., \& Kellman, P. J. (1994).
	\newblock Spatiotemporal boundary formation: Boundary, form, and motion perception from transformations of surface elements.
	\newblock \emph{Journal of Experimental Psychology: General}, 123(1):3.
	
	\bibitem[{Von Der Heydt et~al.(1993)Von Der Heydt, R., Heitger, F., \& Peterhans, E.}]{von1993perception}
	Von Der Heydt, R., Heitger, F., \& Peterhans, E. (1993).
	\newblock Perception of occluding contours: Neural mechanisms and a computational model.
	\newblock \emph{Biomedical research}, 14:1--6.

	\bibitem[{Wagemans et~al.(2012)Wagemans, J., Elder, J. H., Kubovy, M., Palmer, S. E., Peterson, M. A., Singh, M., \& von der Heydt, R.}]{wagemans2012century}
Wagemans, J., Elder, J. H., Kubovy, M., Palmer, S. E., Peterson, M. A., Singh, M., \& von der Heydt, R.(2012).
	\newblock  Century of Gestalt Psychology in Visual Perception: I. Perceptual grouping and Figure-Ground Organization.
	\newblock \emph{Psychological bulletin}, 138(6):1172.		
	
	\bibitem[{Weiss(1999)}]{weiss1999segmentation}
	Weiss, Y. (1999).
	\newblock Segmentation using eigenvectors: a unifying view.
	\newblock \emph{Computer vision, 1999. The proceedings of the seventh IEEE international conference on}, (2):975--982.

	\bibitem[{Wertheimer(1938).}]{wertheimer1938laws}
	Wertheimer, M. (1938).
	\newblock Laws of organization in perceptual forms.
	\newblock \emph{London: Harcourt (Brace and Jovanovich)}.

	\bibitem[{Williams \& Jacobs(1997)Williams, L. R., \& Jacobs, D. W. }]{williams1997stochastic}
	Williams, L. R., \& Jacobs, D. W. (1997).
	\newblock Stochastic completion fields.
	\newblock \emph{Neural computation}, 9(4):837--858.		
	
	\bibitem[{Zucker(2006)}]{zucker2006differential}
	Zucker, S. (2006)
	\newblock Differential geometry from the Frenet point of view: boundary detection, stereo, texture and color.
	\newblock \emph{Handbook of Mathematical Models in Computer Vision}, 357--373, Springer.

	
\end{thebibliography}
\end{document}